\documentclass[letterpaper]{article} 
\usepackage[preprint]{aaai2027}  
\usepackage[hyphens]{url}  
\usepackage{graphicx} 
\usepackage{natbib}  
\usepackage{caption} 
\usepackage{multirow}
\usepackage{amssymb}
\usepackage{amsmath}
\usepackage{algorithm}
\usepackage{algorithmic}
\usepackage{booktabs}

\title{ReDiff: Reliability-Guided Diffusion for Trustworthy Ultra-Low-Field to High-Field MRI Synthesis}

\author{
    Zhenxuan Zhang\textsuperscript{\rm 1},
    Peiyuan Jing\textsuperscript{\rm 1},
    Ruicheng Yuan\textsuperscript{\rm 1},
    Liwei Hu\textsuperscript{\rm 1},
    Anbang Wang\textsuperscript{\rm 1},
    Fanwen Wang\textsuperscript{\rm 1},
    Yinzhe Wu\textsuperscript{\rm 1},
    Kh Tohidul Islam\textsuperscript{\rm 5},
    Zhaolin Chen\textsuperscript{\rm 5},
    Zi Wang\textsuperscript{\rm 1},
    Peter Lally\textsuperscript{\rm 1},
    Guang Yang\textsuperscript{\rm 1,2,3,4}
}
\affiliations{
    \textsuperscript{\rm 1}Department of Bioengineering and Imperial-X, Imperial College London, London, UK\\
    \textsuperscript{\rm 2}National Heart and Lung Institute, Imperial College London, London SW7 2AZ, UK\\
    \textsuperscript{\rm 3}Cardiovascular Research Centre, Royal Brompton Hospital, London SW3 6NP, UK\\
    \textsuperscript{\rm 4}School of Biomedical Engineering \& Imaging Sciences, King's College London, London WC2R 2LS, UK\\
    \textsuperscript{\rm 5}Department of Electrical and Computer Systems Engineering, Monash University, Melbourne, Australia\\
    \{zz2524, peiyuan.jing22, r.yuan25, l.hu25, w.anbang25, fanwen.wang, yinzhe.wu18, zi.wang, p.lally, g.yang\}@imperial.ac.uk,
    \{khtohidul.islam, zhaolin.chen\}@monash.edu
}

\begin{document}

\maketitle

\begin{abstract}
Low-field to high-field MRI synthesis has emerged as a promising strategy to improve image quality when access to high-field scanners is limited. However, in ultra-low-field settings, the degradation of anatomical detail is spatially heterogeneous: structurally ambiguous regions are more susceptible to unstable high-frequency generation, which may produce anatomically inconsistent textures and boundaries. This issue is particularly problematic when synthesized images are used for downstream quantitative analysis. We therefore study how to make diffusion-based LF-to-HF synthesis more spatially reliable, rather than only sharper on average. To this end, we propose a reliability-guided diffusion framework (ReDiff) with two complementary inference-time mechanisms. First, a reliability-guided sampling strategy attenuates unstable reverse-diffusion updates in regions with weak low-field support. Second, an uncertainty-aware candidate selection scheme aggregates multiple stochastic reconstructions according to spatial consensus and predictive uncertainty. Beyond aggregate image quality, we test whether the uncertainty is itself a usable reliability signal. Experiments on paired 64mT$\rightarrow$3T MRI datasets show that ReDiff attains the lowest LPIPS across three contrasts and two datasets while remaining competitive on PSNR and SSIM, and downstream segmentation analysis indicates better preservation of anatomical structure. 
\end{abstract}


\section{Introduction}
High-field (HF) MRI systems (e.g., 3T) provide superior signal-to-noise ratio and improved depiction of fine anatomical structures, particularly along tissue boundaries and cortical folding patterns \cite{arnold2023low,brown2014magnetic}. They are also more reliable in resolving low-contrast soft-tissue regions and small pathological variations. In contrast, low-field (LF) MRI (e.g., 64 mT) typically exhibits reduced spatial resolution, elevated noise levels, and blurred structural boundaries, with degradation being most pronounced in thin cortical regions, deep gray matter structures and other anatomically complex areas \cite{arnold2023low,mazurek2021portable}. As a result, the quality gap between LF and HF imaging is highly spatially heterogeneous rather than uniform across the image \cite{Islam2025,Yang2025}. Therefore, LF-to-HF MRI synthesis has emerged as a promising computational strategy to reduce this spatially heterogeneous quality gap and enhance structural fidelity without requiring hardware upgrades \cite{Man2023}. For clinical adoption, visual realism alone is insufficient: synthesized images must also be anatomically faithful and quantitatively reliable.

\begin{figure}[t!]
    \centering
    \includegraphics[width=\columnwidth]{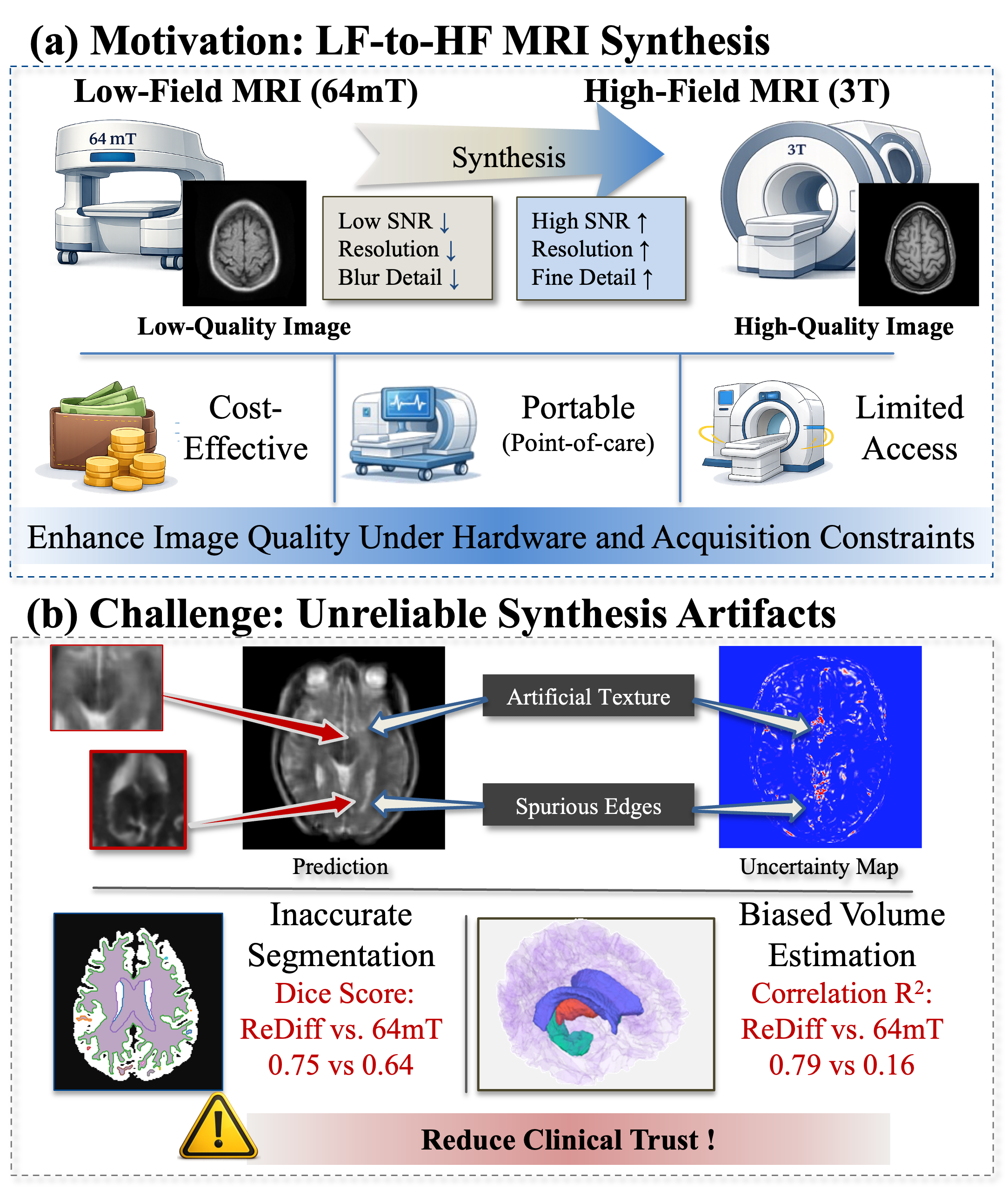}
    \caption{Motivation and challenges of our Reliability-Guided Diffusion.
    (A) Low-field MRI (e.g., 64 mT) is cost-effective and accessible but typically produces lower-quality images. LF-to-HF synthesis seeks to recover high-field (e.g., 3T) image fidelity under hardware and acquisition constraints.
    (B) Despite recent progress, diffusion-based methods may introduce unreliable high-resolution artifacts, such as artificial textures and spurious edges. These errors can degrade downstream tasks, leading to inaccurate segmentation, biased volume estimation, and reduced clinical trust.}
    \label{fig:motivation}
\end{figure}

Despite recent advances, LF-to-HF synthesis remains challenged by unreliable artifacts \cite{Islam2025,Yang2025,Su2024,Ding2026,MiDiffusion,SynDiff}. These arise when conditional generators over-amplify detail in regions where the low-field observation provides only weak structural evidence, and manifest as spurious edges, artificial textures, or anatomically implausible local patterns (Fig.~\ref{fig:motivation}B) \cite{Yang2025,Javadi2025}. Such failures concentrate in anatomically complex regions, thin tissue boundaries, and low-contrast structures where posterior uncertainty is inherently higher \cite{MiDiffusion,SynDiff}, which makes them difficult to control with globally uniform objectives, and even subtle local inconsistencies may propagate to downstream analyses.

\begin{figure*}[t]
\centering
\includegraphics[width=\textwidth]{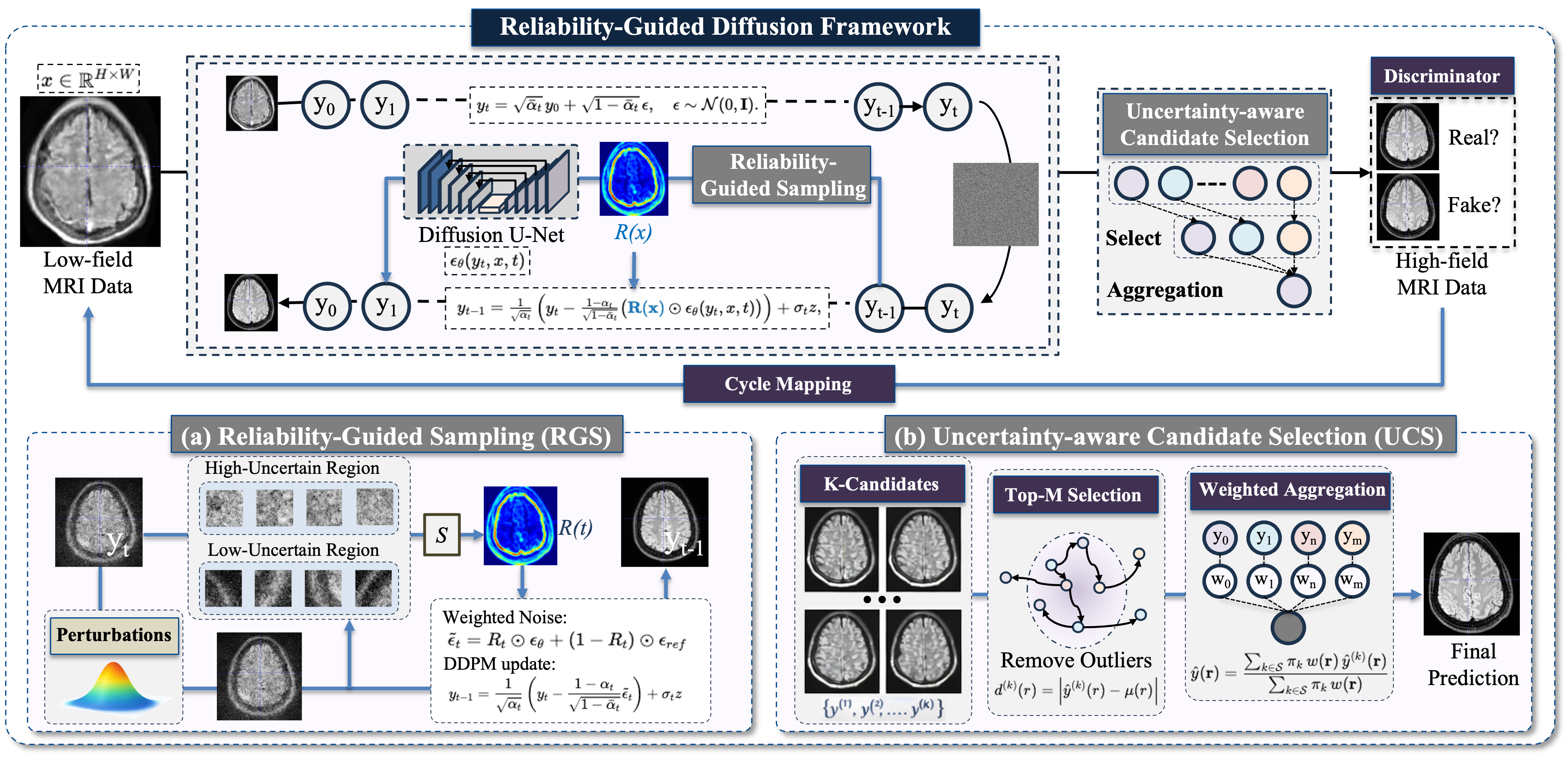}
\caption{Reliability-Guided Diffusion (ReDiff) framework for LF-to-HF MRI synthesis. Given a low-field (LF) image $x$, a conditional diffusion U-Net predicts the noise term $\epsilon_\theta(y_t,x,t)$ to iteratively recover the high-field (HF) image.
(a) In the proposed reliability-guided sampling (RGS), a timestep-dependent reliability map $R_t(x,y_t)$ modulates the reverse diffusion update to suppress unsupported high-frequency amplification in ill-posed regions.
(b) To further enhance robustness, uncertainty-aware candidate selection (UCS) generates multiple candidates and performs reliability-weighted aggregation based on spatial uncertainty.
Together, the two-stage design improves structural consistency while reducing hallucinated details.
}
\label{fig:rediff_framework}
\end{figure*}

Diffusion models recover high-resolution detail and improve perceptual quality for this task \cite{ho2020denoising,zhang2025pretext,zhang2025cyclic}, yet they optimize and decode in a largely spatially uniform manner and so do not distinguish regions strongly supported by the low-field observation from intrinsically ambiguous ones \cite{zhang2025coarse,Dayarathna2025}. Perceptual losses, adversarial regularization, and multi-scale architectures improve sharpness without answering the reliability question that matters here: when the input evidence is weak, how should the generator avoid overconfident high-frequency hallucination while still recovering anatomically meaningful detail? This motivates a conditional generator that models where synthesis is likely to be reliable and where it should be conservative.

To address these limitations, we propose a reliability-aware diffusion framework, ReDiff, that introduces explicit spatial reliability control at inference time. Our approach integrates two complementary mechanisms. First, reliability-guided sampling (RGS) modulates the reverse diffusion update using a sensitivity-derived reliability estimate, thereby suppressing unstable denoising responses in ill-posed regions. Second, uncertainty-aware candidate selection (UCS) aggregates multiple stochastic reconstructions according to spatial consensus and predictive uncertainty, reducing the impact of outlier samples. Rather than framing the method only as a sharper generator, we formulate it as a more conservative conditional synthesis strategy for anatomically ambiguous regions. We evaluate this design through paired LF-to-HF MRI synthesis, ablations, and downstream anatomical analysis. Our contributions are summarized as follows:
\begin{itemize}
    \item We identify spatially unreliable high-frequency generation as a central failure mode in LF-to-HF MRI synthesis and motivate reliability-aware conditional generation for anatomically ambiguous regions.
    \item We propose ReDiff, a diffusion-based synthesis framework that combines reliability-guided sampling with uncertainty-aware candidate selection to improve inference-time robustness.
    \item We show that the uncertainty estimated by UCS functions as a reliability signal rather than a by-product: it stratifies slice-level fidelity monotonically and separates low-fidelity cases with a roughly sixfold difference in failure rate between the most and least uncertain strata.
    \item We evaluate ReDiff on paired 64mT-to-3T MRI through quantitative comparison against GAN-, transformer-, and diffusion-based baselines, module ablation, inference-cost analysis, and downstream segmentation and volumetric assessment.
\end{itemize}

\section{Related Work}
\subsubsection{LF-to-HF MRI Synthesis.}
Paired translation, multimodal synthesis, and task-adapted reconstruction have all been applied to LF-to-HF MRI \cite{isola2017image,zhu2017cyclegan,zhang2025pretext,zhang2025cyclic,liu2024survey}, with later work raising capacity through transformers, cross-modality priors, and consistency-aware objectives \cite{transunet,resvit,cytran,MiDiffusion}. These methods narrow the visual gap between field strengths, yet they optimize global reconstruction quality and leave generator behaviour unregulated where LF evidence is weak or structurally ambiguous \cite{Dayarathna2025,liu2024survey}.
 
\subsubsection{Diffusion-based Medical Image Translation.}
Diffusion models now lead medical image translation, modelling complex conditional distributions and recovering sharper detail than deterministic generators \cite{ho2020denoising,song2021ddim,saharia2022palette,SynDiff}. Adversarial diffusion, mutual-information guidance, cyclic constraints, and score-based priors for MRI inverse problems have improved realism and cross-domain consistency \cite{SynDiff,MiDiffusion,chung2022score,zhang2025cyclic,zhang2025coarse}. These samplers nevertheless remain spatially uniform: once conditioned, every location shares the same denoising dynamics. They therefore do not address the failure mode central to ultra-low-field synthesis, where posterior uncertainty varies sharply across anatomy and can trigger locally unstable high-frequency generation \cite{Yang2025,Javadi2025}.
 
\subsubsection{Reliability and Uncertainty in Medical Generation.}
Uncertainty in medical reconstruction and enhancement has been estimated through Bayesian approximation, ensembles, and variational posteriors \cite{gal2016dropout,kendall2017uncertainties,lakshminarayanan2017simple,schlemper2018bayesian,edupuganti2021uncertainty,narnhofer2022bayesian}, and used to flag unsafe neuroimage enhancement or to steer acquisition \cite{tanno2021uncertainty,zhang2019reducing}; in synthesis, visually plausible outputs can still be locally unreliable \cite{Dayarathna2025,Yang2025,Javadi2025}. In nearly all of this work uncertainty is an analysis product or a global auxiliary constraint. We instead place reliability control inside conditional diffusion inference, so that local sensitivity modulates the denoising trajectory and uncertainty-aware aggregation governs candidate selection.
\begin{figure*}[t]
\centering
\includegraphics[width=\textwidth]{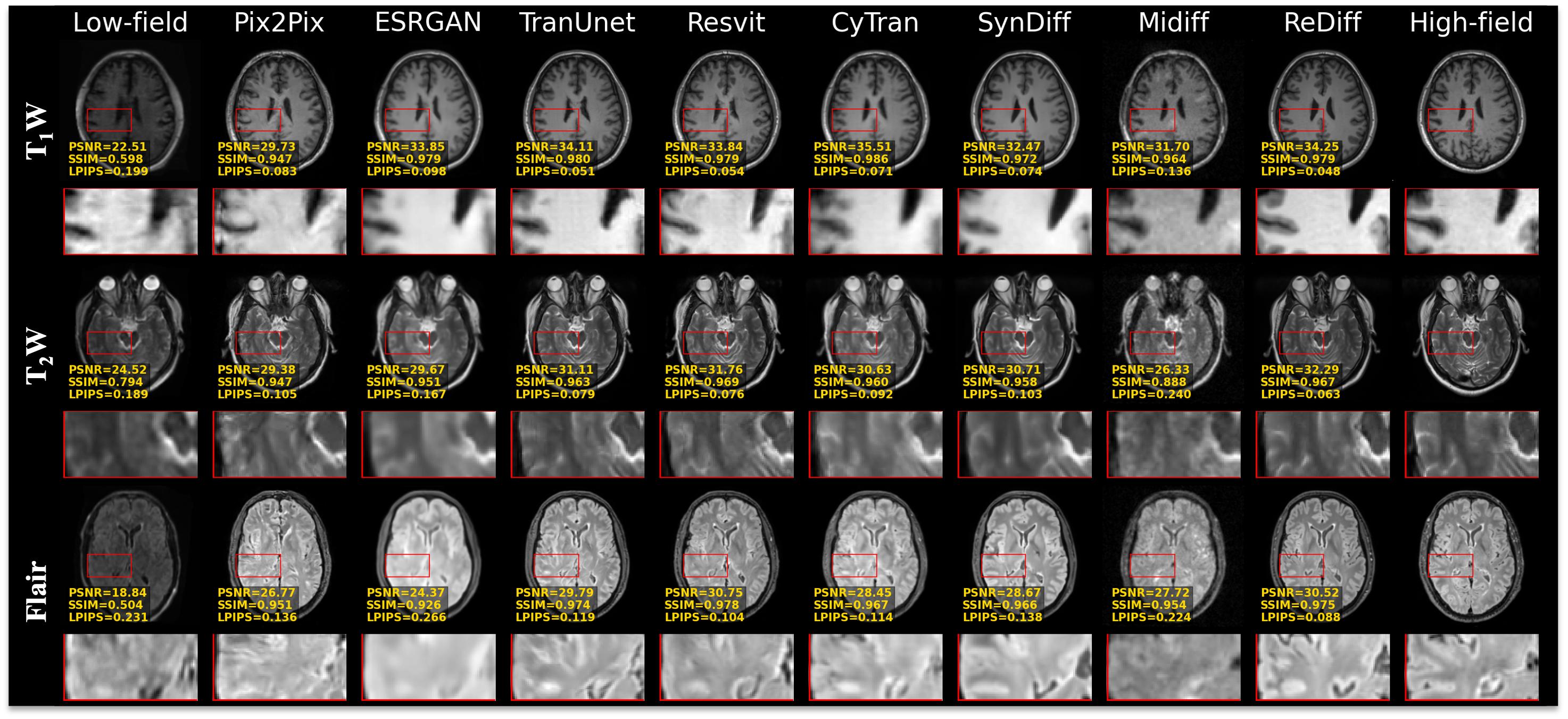}
\caption{
Qualitative comparison of multi-contrast low-field to high-field MRI synthesis. For each case, the top row shows full images and the bottom row shows zoomed views of the red boxed regions. ReDiff produces sharper anatomical structures and fewer artifacts, with PSNR and SSIM reported on each result.
}
\label{fig:compare}
\end{figure*}

\section{Methodology}

\subsubsection{Problem Formulation.} Let $y \in \mathbb{R}^{H\times W\times D}$ denote the latent HF MRI volume acquired at main field strength $B_H$ and $x \in \mathbb{R}^{H\times W\times D}$ the observed LF volume acquired at $B_L$ ($B_L < B_H$). We state the formulation volumetrically for clarity; the generator itself operates on co-registered 2D axial slices, and volumes are reassembled slice-wise before downstream analysis.
We model the LF image as a field-strength-dependent observation of the same underlying anatomy:
\begin{equation}
x = \mathcal{F}_{B_L}(y) + \epsilon_{B_L},
\qquad
\epsilon_{B_L}\sim \mathcal{N}\!\left(0,\sigma^2(B_L)\mathbf I\right),
\end{equation}
where $\mathcal{F}_{B_L}$ represents the effective low-field imaging operator that attenuates high-frequency information and modifies tissue contrast, and $\sigma^2(B_L)$ reflects the increased noise level associated with reduced $B_0$. Due to the loss of high-frequency observability and elevated noise at low field, recovering $y$ from $x$ is intrinsically ill-posed. This uncertainty makes conditional generation prone to spatially unreliable high-frequency hallucinations in regions where the LF evidence is weak.

Diffusion-based approaches address this task by learning the conditional distribution $p_\theta(y\mid x)$ and generating $\hat{y}$ via iterative denoising. However, in regions where $\mathcal{F}_{B_L}$ severely attenuates high-frequency information, the posterior becomes highly uncertain, which may lead to over-amplified high-frequency responses and spatially inconsistent structures during sampling. To mitigate this issue, we seek a reliability-aware generator
\begin{equation}
\hat{y} = \mathcal{G}(x;\theta,\phi),
\end{equation}
where $\theta$ denotes the diffusion network parameters and $\phi$ denotes inference-time reliability-control parameters. The goal of $\mathcal{G}$ is not to maximize sharpness uniformly across the image, but to preserve anatomically supported detail while attenuating unstable updates in regions with weak low-field evidence.

\subsubsection{Conditional Diffusion Backbone.}
Our starting point is a conditional diffusion model that learns the reverse transition from noisy HF variables to the clean HF target under LF conditioning. Given a clean HF target $y_0$, the forward diffusion process perturbs it into
\begin{equation}
y_t=\sqrt{\bar{\alpha}_t}y_0+\sqrt{1-\bar{\alpha}_t}\epsilon,\qquad \epsilon\sim\mathcal N(0,\mathbf I),
\label{eq:forward}
\end{equation}
and a conditional U-Net predicts the injected noise through $\epsilon_\theta(y_t,x,t)$. In a standard conditional diffusion sampler, the predicted noise is directly used in each reverse step. ReDiff keeps this backbone unchanged and instead introduces reliability control during inference, so that the denoising dynamics can adapt to the spatial confidence implied by the LF input.

\subsubsection{Reliability-Guided Sampling (RGS).} Following the standard formulation \cite{ho2020denoising}, the reverse diffusion step is
\begin{equation}
\begin{aligned}
y_{t-1} = \frac{1}{\sqrt{\alpha_t}} \left( y_t - \frac{1-\alpha_t}{\sqrt{1-\bar{\alpha}_t}} \,\epsilon_\theta(y_t, x, t) \right) + \sigma_t z, \\
z\sim\mathcal{N}(0,\mathbf I),
\end{aligned}
\end{equation}
where $\epsilon_\theta(\cdot)$ is the predicted noise on the LF image $x$.
 
Where LF evidence is weak the posterior broadens and this update may over-amplify high-frequency responses. Our key observation is that the resulting instability can be probed through the local sensitivity of the noise predictor: if a small perturbation of the noisy state changes the predicted noise substantially, the current update is less trustworthy at that location. This yields an operational surrogate for spatial reliability that needs no additional supervision. We inject $J$ Gaussian perturbations $\delta_j\sim\mathcal{N}(0,\sigma_p^2 \mathbf I)$ into $y_t$ and take
\begin{equation}
\tilde{S}_t = \frac{1}{J}\sum_{j=1}^{J}\left|\epsilon_\theta(y_t+\delta_j,x,t)- \epsilon_\theta(y_t,x,t)\right|,
\end{equation}
followed by the timestep-dependent reliability map $R_t(x,y_t)\in[R_{\min},1]^{H\times W\times D}$,
\begin{equation}
R_t(x,y_t)=R_{\min}+(1-R_{\min})\exp(-\gamma \tilde{S}_t),
\end{equation}
where $\gamma$ controls the attenuation strength and the floor $R_{\min}\in(0,1)$ bounds how far any location may depart from the standard sampler. The noise estimate is then replaced by a reliability-weighted interpolation between the network prediction and a conservative reference $\epsilon_{\mathrm{ref}}$,
\begin{equation}
\tilde{\epsilon}_t = R_t(x,y_t)\odot\epsilon_\theta(y_t,x,t) + \big(\mathbf 1-R_t(x,y_t)\big)\odot\epsilon_{\mathrm{ref}},
\end{equation}
and the reverse step is taken with $\tilde{\epsilon}_t$ in place of $\epsilon_\theta$:
\begin{equation}
y_{t-1}=\frac{1}{\sqrt{\alpha_t}}\left(
y_t-\frac{1-\alpha_t}{\sqrt{1-\bar{\alpha}_t}}\,\tilde{\epsilon}_t
\right)+\sigma_t z,
\label{eq:rgs_update}
\end{equation}
where $\odot$ is element-wise multiplication and $\mathbf 1$ the all-ones tensor. We set $\epsilon_{\mathrm{ref}}=0$ throughout, which recovers pure attenuation of the denoising correction; the interpolation form keeps the reference explicit and admits alternatives such as a low-pass filtered conditioning path. RGS modifies only the noise estimate consumed at each reverse step, leaving the backbone unchanged.
 
Eq.~\eqref{eq:rgs_update} is not the exact reverse transition of Eq.~\eqref{eq:forward}, and we do not claim that it preserves the marginals of $p_\theta(y\mid x)$. Attenuating $\epsilon_\theta$ below unity leaves a residual fraction of $y_t$ in $y_{t-1}$, shrinking the trajectory toward the conditional low-frequency content with a per-step bias of order $\tfrac{1-\alpha_t}{\sqrt{1-\bar\alpha_t}}\lVert(\mathbf 1-R_t)\odot\epsilon_\theta\rVert$. This bias is the intended effect rather than an approximation error: it trades high-frequency energy in unstable regions for reduced variance across stochastic runs, and $\gamma$ together with $R_{\min}$ bounds its magnitude so that well-supported anatomy is left essentially untouched.

\begin{table*}[t!]
\centering
\caption{Evaluation on the paired 64mT$\rightarrow$3T dataset under Multi-Contrast setting. $^{*}$ indicates $p < 0.05$ and $^{**}$ indicates $p < 0.01$ in a paired Wilcoxon signed-rank test against our ReDiff model, computed over test slices; absence of a marker indicates that the difference from ReDiff is not significant. \textbf{Bold} indicates the best result and \underline{underline} indicates the second best.}
\label{tab:comparison}
\resizebox{\textwidth}{!}{
\begin{tabular}{l|l|ccc|ccc}
\hline\hline
\multirow{2}{*}{\textbf{Setting}} & \multirow{2}{*}{\textbf{Method}} & \multicolumn{3}{c|}{\textbf{Private Dataset}} & \multicolumn{3}{c}{\textbf{Leiden Uni. Dataset}} \\
\cline{3-8}
 & & PSNR$\uparrow$ & SSIM$\uparrow$ & LPIPS$\downarrow$ & PSNR$\uparrow$ & SSIM$\uparrow$ & LPIPS$\downarrow$ \\
\hline

\multirow{9}{*}{\textbf{T$_1$w}}
&Low-field
& 24.83$\pm$2.77$^{**}$ & 0.758$\pm$0.133$^{**}$ & 0.2222$\pm$0.0866$^{**}$
& 24.73$\pm$2.81$^{**}$ & 0.768$\pm$0.117$^{**}$ & 0.2138$\pm$0.0741$^{**}$ \\
& Pix2Pix\cite{isola2017image}
& 30.51$\pm$2.62$^{**}$ & 0.933$\pm$0.052$^{**}$ & 0.1050$\pm$0.0394$^{**}$
& 30.71$\pm$2.64$^{**}$ & 0.939$\pm$0.054$^{**}$ & 0.1043$\pm$0.0366$^{**}$ \\
& ESRGAN\cite{esrgan}
& 31.72$\pm$2.54$^{**}$ & 0.949$\pm$0.038$^{*}$ & 0.1080$\pm$0.0746$^{**}$
& 32.08$\pm$2.46$^{**}$ & 0.958$\pm$0.029 & 0.1034$\pm$0.0704$^{**}$ \\
& TransUNet\cite{transunet}
& 31.71$\pm$2.75$^{**}$ & 0.939$\pm$0.092$^{**}$ & 0.0947$\pm$0.0720$^{*}$
& 32.08$\pm$2.72$^{**}$ & 0.950$\pm$0.069$^{*}$ & 0.0901$\pm$0.0592$^{*}$ \\
& ResViT\cite{resvit}
& 31.61$\pm$2.75$^{**}$ & 0.941$\pm$0.058$^{**}$ & \underline{0.0887$\pm$0.0505}$^{*}$
& 32.10$\pm$2.65$^{**}$ & 0.953$\pm$0.045$^{*}$ & \underline{0.0838$\pm$0.0435} \\
& CyTran\cite{cytran}
& 32.76$\pm$2.78$^{*}$ & \textbf{0.961$\pm$0.065} & 0.0950$\pm$0.0451$^{*}$
& \underline{33.21$\pm$2.65} & \textbf{0.970$\pm$0.047} & 0.0909$\pm$0.0389$^{*}$ \\
& SynDiff\cite{SynDiff}
& \textbf{33.34$\pm$2.36} & 0.953$\pm$0.003 & 0.1102$\pm$0.0361$^{**}$
& 33.20$\pm$2.10 & 0.953$\pm$0.003 & 0.1174$\pm$0.0404$^{**}$ \\
& MiDiffusion\cite{MiDiffusion}
& 32.77$\pm$4.79$^{*}$ & 0.944$\pm$0.053$^{**}$ & 0.1717$\pm$0.0428$^{**}$
& 32.54$\pm$4.78$^{**}$ & 0.942$\pm$0.056$^{**}$ & 0.1714$\pm$0.0410$^{**}$ \\
& \textbf{ReDiff}
& \underline{33.22$\pm$2.71} & \underline{0.960$\pm$0.042} & \textbf{0.0800$\pm$0.0317}
& \textbf{33.70$\pm$2.65} & \underline{0.968$\pm$0.029} & \textbf{0.0774$\pm$0.0282} \\
\hline

\multirow{9}{*}{\textbf{T$_2$w}}
&Low-field
& 24.90$\pm$2.93$^{**}$ & 0.753$\pm$0.132$^{**}$ & 0.2192$\pm$0.0820$^{**}$
& 25.01$\pm$2.94$^{**}$ & 0.765$\pm$0.130$^{**}$ & 0.2142$\pm$0.0830$^{**}$ \\
& Pix2Pix\cite{isola2017image}
& 30.64$\pm$2.71$^{**}$ & 0.932$\pm$0.052$^{**}$ & 0.1072$\pm$0.0381$^{**}$
& 30.78$\pm$2.90$^{**}$ & 0.936$\pm$0.051$^{**}$ & 0.1049$\pm$0.0414$^{**}$ \\
& ESRGAN\cite{esrgan}
& 31.94$\pm$2.59$^{**}$ & 0.950$\pm$0.040$^{*}$ & 0.1069$\pm$0.0770$^{**}$
& 32.22$\pm$2.66$^{*}$ & 0.956$\pm$0.031 & 0.1050$\pm$0.0824$^{**}$ \\
& TransUNet\cite{transunet}
& 31.83$\pm$2.78$^{**}$ & 0.939$\pm$0.091$^{**}$ & 0.0950$\pm$0.0705$^{*}$
& 32.21$\pm$2.92$^{*}$ & 0.945$\pm$0.098$^{*}$ & 0.0932$\pm$0.0774$^{*}$ \\
& ResViT\cite{resvit}
& 31.72$\pm$2.81$^{**}$ & 0.939$\pm$0.065$^{**}$ & \underline{0.0899$\pm$0.0559}$^{*}$
& 32.14$\pm$2.92$^{*}$ & 0.948$\pm$0.064$^{*}$ & \underline{0.0860$\pm$0.0576} \\
& CyTran\cite{cytran}
& 32.93$\pm$2.88$^{*}$ & \textbf{0.963$\pm$0.049} & 0.0952$\pm$0.0455$^{*}$
& 33.23$\pm$3.01 & \textbf{0.967$\pm$0.051} & 0.0962$\pm$0.0505$^{*}$ \\
& SynDiff\cite{SynDiff}
& \underline{33.17$\pm$2.39} & 0.952$\pm$0.004 & 0.1150$\pm$0.0399$^{**}$
& \underline{33.32$\pm$2.39} & 0.953$\pm$0.003 & 0.1148$\pm$0.0383$^{**}$ \\
& MiDiffusion\cite{MiDiffusion}
& 32.11$\pm$4.83$^{**}$ & 0.931$\pm$0.063$^{**}$ & 0.1787$\pm$0.0438$^{**}$
& 32.57$\pm$4.80$^{*}$ & 0.940$\pm$0.059$^{**}$ & 0.1740$\pm$0.0436$^{**}$ \\
& \textbf{ReDiff}
& \textbf{33.38$\pm$2.66} & \underline{0.961$\pm$0.034} & \textbf{0.0804$\pm$0.0319}
& \textbf{33.66$\pm$2.86} & \underline{0.964$\pm$0.040} & \textbf{0.0800$\pm$0.0372} \\
\hline

\multirow{9}{*}{\textbf{FLAIR}}
&Low-field
& 24.68$\pm$2.85$^{**}$ & 0.758$\pm$0.115$^{**}$ & 0.2238$\pm$0.0856$^{**}$
& 24.98$\pm$2.95$^{**}$ & 0.757$\pm$0.114$^{**}$ & 0.2215$\pm$0.0828$^{**}$ \\
& Pix2Pix\cite{isola2017image}
& 30.64$\pm$2.51$^{**}$ & 0.937$\pm$0.040$^{**}$ & 0.1079$\pm$0.0423$^{**}$
& 30.97$\pm$2.61$^{**}$ & 0.938$\pm$0.048$^{**}$ & 0.1048$\pm$0.0334$^{**}$ \\
& ESRGAN\cite{esrgan}
& 31.86$\pm$2.52$^{**}$ & 0.953$\pm$0.034$^{*}$ & 0.1139$\pm$0.0940$^{**}$
& 32.08$\pm$2.75$^{*}$ & 0.951$\pm$0.041$^{*}$ & 0.1074$\pm$0.0759$^{**}$ \\
& TransUNet\cite{transunet}
& 31.88$\pm$2.71$^{**}$ & 0.948$\pm$0.051$^{*}$ & 0.0973$\pm$0.0745$^{*}$
& 32.08$\pm$2.86$^{*}$ & 0.945$\pm$0.057$^{*}$ & 0.0943$\pm$0.0557$^{*}$ \\
& ResViT\cite{resvit}
& 31.94$\pm$2.68$^{**}$ & 0.949$\pm$0.048$^{*}$ & \underline{0.0912$\pm$0.0750}$^{*}$
& 32.24$\pm$2.79$^{*}$ & 0.949$\pm$0.045$^{*}$ & \underline{0.0865$\pm$0.0480} \\
& CyTran\cite{cytran}
& \underline{33.05$\pm$2.72}$^{*}$ & \textbf{0.969$\pm$0.031} & 0.0996$\pm$0.0660$^{*}$
& \underline{33.31$\pm$2.94} & \textbf{0.966$\pm$0.046} & 0.0941$\pm$0.0424$^{*}$ \\
& SynDiff\cite{SynDiff}
& 32.96$\pm$2.12 & 0.953$\pm$0.005 & 0.1169$\pm$0.0420$^{**}$
& 33.13$\pm$2.35 & 0.953$\pm$0.003 & 0.1178$\pm$0.0385$^{**}$ \\
& MiDiffusion\cite{MiDiffusion}
& 32.23$\pm$4.66$^{**}$ & 0.936$\pm$0.061$^{**}$ & 0.1792$\pm$0.0578$^{**}$
& 32.62$\pm$4.62$^{*}$ & 0.939$\pm$0.057$^{**}$ & 0.1773$\pm$0.0435$^{**}$ \\
& \textbf{ReDiff}
& \textbf{33.47$\pm$2.62} & \underline{0.964$\pm$0.038} & \textbf{0.0849$\pm$0.0552}
& \textbf{33.63$\pm$2.97} & \underline{0.957$\pm$0.078} & \textbf{0.0828$\pm$0.0485} \\
\hline\hline
\end{tabular}}
\end{table*}

\begin{table}[t]
\centering
\caption{Ablation study of the proposed ReDiff framework on the Private and Leiden datasets, averaged over the T$_1$w, T$_2$w, and FLAIR contrasts. RGS denotes reliability-guided sampling and UCS denotes uncertainty-aware candidate selection. Bold indicates the full model.}
\label{tab:ablation}
\resizebox{\columnwidth}{!}{
\begin{tabular}{c c c | ccc}
\hline\hline
\multirow{2}{*}{\textbf{Diff}} & \multirow{2}{*}{\textbf{RGS}} & \multirow{2}{*}{\textbf{UCS}}
& \multicolumn{3}{c}{\textbf{Private Dataset}} \\
\cline{4-6}
& & & PSNR$\uparrow$ & SSIM$\uparrow$ & LPIPS$\downarrow$ \\
\hline
$\checkmark$ & $\times$ & $\times$
& 30.98$\pm$2.55 & 0.934$\pm$0.060 & 0.1080$\pm$0.0697 \\
$\checkmark$ & $\checkmark$ & $\times$
& 31.72$\pm$2.51 & 0.945$\pm$0.052 & 0.1062$\pm$0.0717 \\
$\checkmark$ & $\times$ & $\checkmark$
& 31.45$\pm$2.60 & 0.940$\pm$0.056 & 0.0991$\pm$0.0687 \\
\hline
$\checkmark$ & $\checkmark$ & $\checkmark$
& \textbf{33.00$\pm$2.66} & \textbf{0.959$\pm$0.038} & \textbf{0.0838$\pm$0.0595} \\
\hline

\multirow{2}{*}{\textbf{Diff}} & \multirow{2}{*}{\textbf{RGS}} & \multirow{2}{*}{\textbf{UCS}}
& \multicolumn{3}{c}{\textbf{Leiden Uni. Dataset}} \\
\cline{4-6}
& & & PSNR$\uparrow$ & SSIM$\uparrow$ & LPIPS$\downarrow$ \\
\hline
$\checkmark$ & $\times$ & $\times$
& 31.32$\pm$2.63 & 0.940$\pm$0.054 & 0.1038$\pm$0.0628 \\
$\checkmark$ & $\checkmark$ & $\times$
& 32.11$\pm$2.67 & 0.950$\pm$0.049 & 0.1022$\pm$0.0622 \\
$\checkmark$ & $\times$ & $\checkmark$
& 31.82$\pm$2.73 & 0.946$\pm$0.057 & 0.0946$\pm$0.0587 \\
\hline
$\checkmark$ & $\checkmark$ & $\checkmark$
& \textbf{33.30$\pm$2.77} & \textbf{0.961$\pm$0.043} & \textbf{0.0806$\pm$0.0490} \\
\hline\hline
\end{tabular}}
\end{table}

\subsubsection{Uncertainty-aware Candidate Selection (UCS).}
While the proposed reliability-guided sampling reduces the risk of unreliable updates during generation, residual uncertainty may still remain due to the intrinsic ill-posedness of LF-to-HF synthesis. To further improve robustness, we introduce an uncertainty-aware candidate selection scheme as a complementary post-generation safeguard.

Given the LF input $x$, we generate $K$ candidate HF reconstructions by stochastic diffusion sampling:
\begin{equation}
\{\hat{y}^{(k)}\}_{k=1}^{K} \sim p_\theta(y\mid x).
\end{equation}
These candidates reflect the posterior variability induced by noise and ambiguous LF evidence.

To suppress structurally inconsistent samples, we first measure the similarity of each candidate to the consensus. Let
$ \mu(\mathbf r)=\frac{1}{K}\sum_{k=1}^{K}\hat{y}^{(k)}(\mathbf r) $
denote the candidate mean. We compute a deviation score
$ d^{(k)}(\mathbf r)=|\hat{y}^{(k)}(\mathbf r)-\mu(\mathbf r)|. $
Candidates with large deviations are regarded as potential outliers. We retain the top-$M$ most consistent candidates according to the spatially averaged deviation score and denote this filtered set by $\mathcal{K}_M$. On the filtered candidate set, we estimate spatial uncertainty via the empirical variance:
\begin{equation}
U(\mathbf r)
=
\mathrm{Var}\big(\{\hat{y}^{(k)}(\mathbf r)\}_{k\in \mathcal{K}_M}\big).
\end{equation}
The final reconstruction is obtained by reliability-weighted aggregation:
\begin{equation}
\begin{aligned}
\hat{y}(\mathbf r) = \frac{\sum_{k\in \mathcal{K}_M} w^{(k)}(\mathbf r)\,\hat{y}^{(k)}(\mathbf r)} {\sum_{k\in \mathcal{K}_M} w^{(k)}(\mathbf r)}, \\
w^{(k)}(\mathbf r) = \exp\!\Big( -\beta\,U(\mathbf r)\, d^{(k)}(\mathbf r) \Big),
\end{aligned}
\end{equation}
where $\beta$ controls the strength of uncertainty suppression, and both $\mu$ and $d^{(k)}$ are computed against the full candidate mean so that filtering and weighting share a common reference. Since $U(\mathbf r)$ does not depend on $k$, it favors no particular candidate; it acts as a per-voxel temperature on the deviation term. Where the retained candidates disagree, $U(\mathbf r)$ is large and the weights concentrate on those nearest the local consensus, giving a robust consensus selector; where they agree, $U(\mathbf r)$ approaches zero and the weights become nearly uniform, reducing to a plain average that preserves the shared detail. 

\begin{algorithm}[t]
\caption{Reliability-Guided Diffusion Inference}
\label{alg:rediff_algorithm}
\small
\begin{algorithmic}[1]
\STATE \textbf{Input:} Low-field image $x$, trained diffusion predictor $\epsilon_\theta$, diffusion schedule $\{\alpha_t,\bar{\alpha}_t,\sigma_t\}_{t=1}^{T}$, perturbation count $J$ and scale $\sigma_p$, candidate count $K$, retained count $M$, control parameters $\gamma,\beta,R_{\min}$
\STATE \textbf{Output:} Synthesized high-field image $\hat{y}$ and uncertainty map $U$
\FOR{$k=1$ to $K$}
    \STATE Initialize $y_T^{(k)}\sim\mathcal N(0,\mathbf I)$
    \FOR{$t=T$ down to $1$}
        \STATE Predict noise $\epsilon_\theta(y_t^{(k)},x,t)$
        \STATE Estimate local sensitivity: \\
        $\tilde S_t^{(k)}=\frac{1}{J}\sum_{j=1}^{J}\left|\epsilon_\theta(y_t^{(k)}+\delta_j,x,t)-\epsilon_\theta(y_t^{(k)},x,t)\right|$
        \STATE Compute reliability map: \\$R_t^{(k)}=R_{\min}+(1-R_{\min})\exp(-\gamma \tilde S_t^{(k)})$
        \STATE Update $y_{t-1}^{(k)}$ using the RGS reverse step in Eq.~\eqref{eq:rgs_update}
    \ENDFOR
    \STATE Set $\hat y^{(k)}=y_0^{(k)}$
\ENDFOR
\STATE Compute the candidate mean $\mu(\mathbf r)=\frac{1}{K}\sum_{k=1}^{K}\hat y^{(k)}(\mathbf r)$
\STATE Compute deviation scores $d^{(k)}(\mathbf r)=|\hat y^{(k)}(\mathbf r)-\mu(\mathbf r)|$
\STATE Retain the top-$M$ candidates with the smallest spatially averaged deviations to form $\mathcal K_M$
\STATE Estimate spatial uncertainty $U(\mathbf r)=\mathrm{Var}\big(\{\hat y^{(k)}(\mathbf r)\}_{k\in \mathcal K_M}\big)$
\STATE Compute weights $w^{(k)}(\mathbf r)=\exp(-\beta U(\mathbf r)d^{(k)}(\mathbf r))$ for $k\in\mathcal K_M$
\STATE Return $\hat y(\mathbf r)=\frac{\sum_{k\in\mathcal K_M}w^{(k)}(\mathbf r)\hat y^{(k)}(\mathbf r)}{\sum_{k\in\mathcal K_M}w^{(k)}(\mathbf r)}$
\end{algorithmic}
\end{algorithm}

\subsubsection{Training Objective.}
The diffusion backbone is trained with the standard conditional noise-prediction objective together with image-level regularization terms that encourage structural fidelity and perceptual realism. The overall loss is defined as
\begin{equation}
\begin{aligned}
\mathcal{L}
={}& \mathcal{L}_{\mathrm{diff}}
+\lambda_1 \lVert \hat{y} - y \rVert_1
+\lambda_2 \bigl(1-\mathrm{SSIM}(\hat{y},y)\bigr) \\
&+\lambda_3 \mathcal{L}_{\mathrm{adv}}
+\lambda_4 \mathcal{L}_{\mathrm{cyc}} .
\end{aligned}
\end{equation}
where
\begin{equation}
\mathcal{L}_{\mathrm{diff}}=\mathbb{E}_{y,x,t,\epsilon}\left[\|\epsilon-\epsilon_\theta(y_t,x,t)\|_2^2\right]
\end{equation}
is the standard diffusion loss, and $\hat{y}$ and $y$ denote the synthesized and ground-truth HF images, respectively. The $L_1$ and SSIM terms enforce pixel-wise accuracy and structural consistency. $\mathcal{L}_{\mathrm{adv}}$ denotes an image-level adversarial regularizer that encourages realistic high-frequency details, while $\mathcal{L}_{\mathrm{cyc}}$ denotes a cycle-consistency regularizer for stabilizing the LF$\leftrightarrow$HF mapping during training. The hyperparameters $\lambda_1$--$\lambda_4$ balance the auxiliary regularizers.
\begin{figure}[t!]
\centering
\includegraphics[width=\columnwidth]{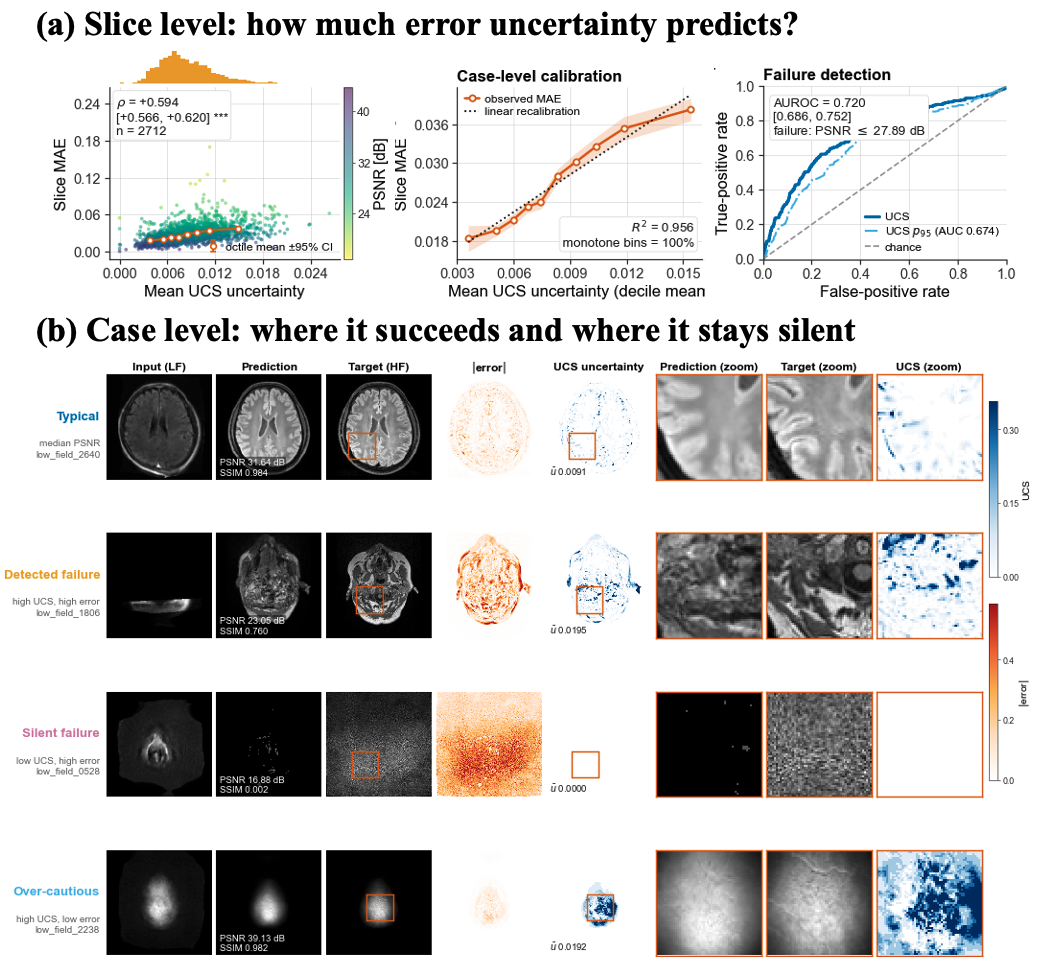}
\caption{
UCS uncertainty as a reliability signal.
(a) Slice-level validation. \textit{Left}: mean UCS uncertainty vs.\ slice MAE,
coloured by PSNR (markers: octile means, 95\% CI). \textit{Centre}: MAE per
uncertainty decile (shaded, 95\% CI) against an affine recalibration (dotted).
\textit{Right}: ROC for identifying bottom-decile slices (PSNR $\leq 27.89$~dB), using mean slice uncertainty as the score.
(b) Representative slices for four diagnostic regimes; $|$error$|$ and UCS share a
colour scale across rows, and the orange box marks the worst error window, magnified at right.
}
\label{fig:ucs_validation}
\end{figure}

\begin{table}[t]
\centering
\caption{Fidelity and failure rate stratified by UCS uncertainty level. Values are mean $\pm$ s.d.; failure rate is reported with its Wilson 95\% confidence interval. $\bar u$ is the mean UCS uncertainty within the stratum.}
\label{tab:ucs-quintiles}
\resizebox{\columnwidth}{!}{
\begin{tabular}{@{}l c c c c c@{}}
\toprule
UCS & $\bar{u}$ & PSNR [dB] & SSIM & Edge MAE & Failure [\%] \\
\midrule
Q1 & 0.0043 & 33.55 $\pm$ 3.16 & 0.9256 $\pm$ 0.1538 & 0.0641 & 4.2 [2.8, 6.3] \\
Q2 & 0.0064 & 32.52 $\pm$ 2.29 & 0.9525 $\pm$ 0.0740 & 0.0702 & 4.1 [2.7, 6.1] \\
Q3 & 0.0079 & 31.63 $\pm$ 2.33 & 0.9573 $\pm$ 0.0386 & 0.0755 & 6.5 [4.7, 8.8] \\
Q4 & 0.0098 & 30.59 $\pm$ 2.50 & 0.9486 $\pm$ 0.0793 & 0.0810 & 10.7 [8.4, 13.6] \\
Q5 & 0.0136 & 29.61 $\pm$ 2.84 & 0.9382 $\pm$ 0.0879 & 0.0884 & 24.7 [21.2, 28.5] \\
\bottomrule
\end{tabular}}
\end{table}

\section{Experiments}
\label{sec:experiments}

\subsubsection{Datasets.}
We evaluate ReDiff on two paired LF--HF MRI datasets: a private set of 20 subjects with T$_1$w, T$_2$w, and FLAIR acquired at both 64\,mT and 3\,T, and the public Leiden set of 11 healthy subjects scanned at both field strengths \cite{leiden_MRI}. Each LF volume is rigidly registered to its 3\,T counterpart, both are resampled to a common grid, and intensities are normalized per volume. 

\subsubsection{Implementation Details.}
ReDiff is trained with Adam using a learning rate of $2\times10^{-5}$, batch size 4, and EMA decay 0.9995. We use 1000 diffusion timesteps with a cosine schedule, $v$-prediction, and 50-step DDIM sampling. UCS uses $K=4$, $\beta=18$, consistency timestep 500, 8 consistency steps, and condition weight 0.15; RGS uses spatial weighting with $\gamma=0.08$ and $R_{\min}=0.94$. All hyperparameters are fixed before testing on Leiden. Experiments use one NVIDIA RTX PRO 6000 Blackwell GPU.

\begin{figure}[t!]
\centering
\includegraphics[width=\columnwidth]{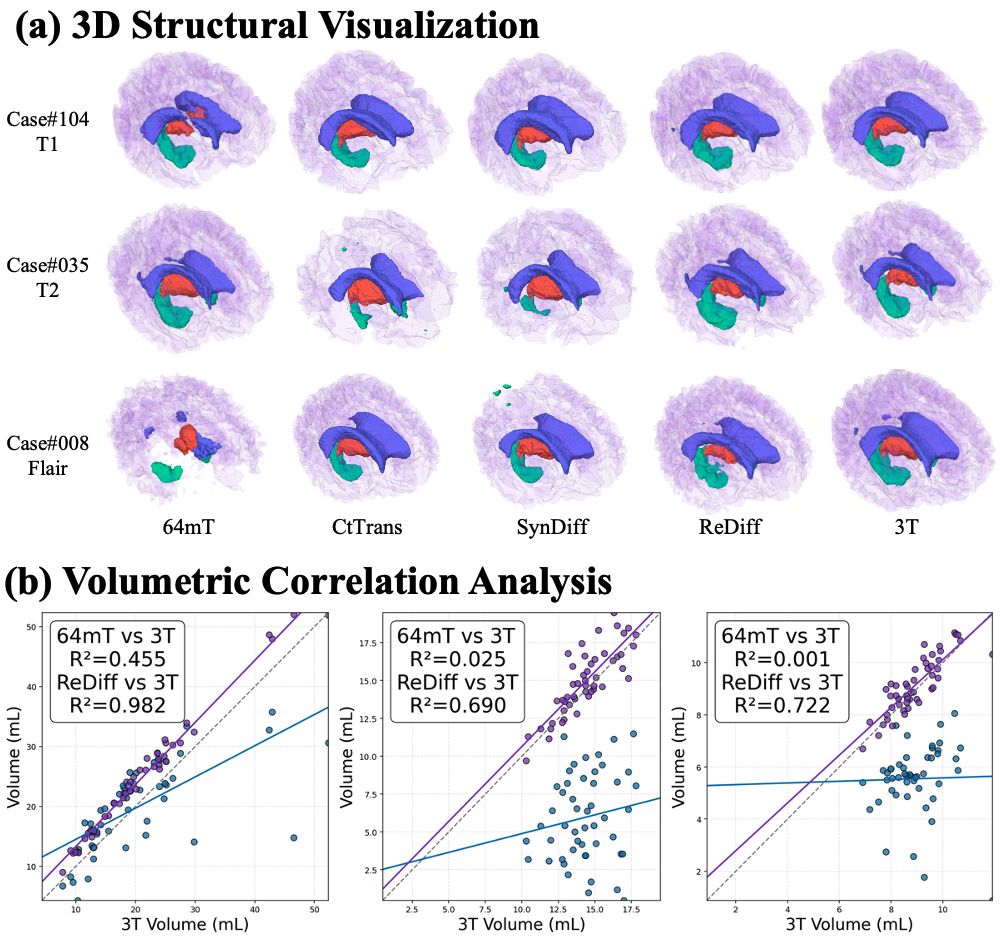}
\caption{
Qualitative and quantitative downstream validation. (a) 3D structural visualization of SynthSeg-derived segmentations for representative cases. (b) Volumetric correlation with respect to the 3T reference.
}
\label{fig:violin}
\end{figure}
\begin{figure}[t]
\centering
\includegraphics[width=\columnwidth]{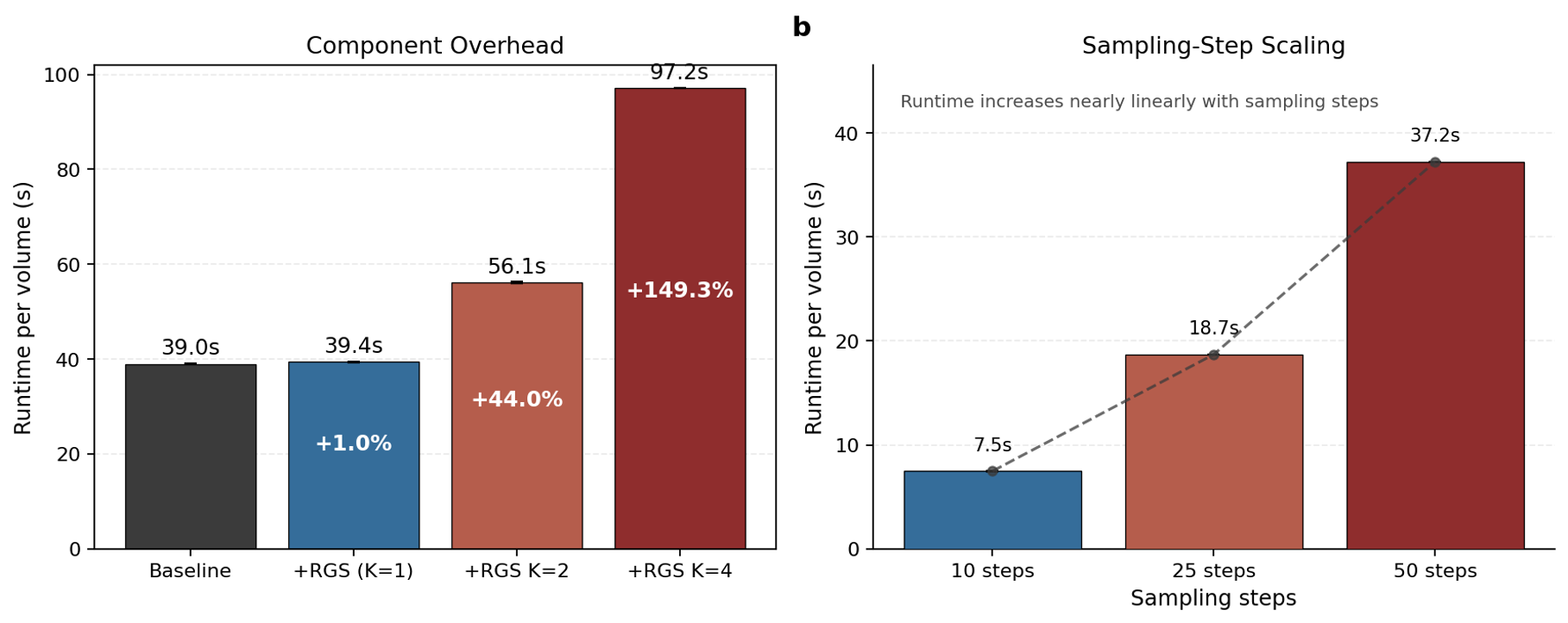}
\caption{
Inference-efficiency analysis of ReDiff. (a) Component overhead measured as runtime per volume under different inference configurations. (b) Sampling-step scaling analysis. Runtime decreases substantially as the number of diffusion sampling steps is reduced.
}
\label{fig:efficiency}
\end{figure}

\subsubsection{Comparison Experiment.}
Table~\ref{tab:comparison} compares methods on the paired 64mT$\rightarrow$3T datasets. ReDiff attains the highest PSNR in five of the six contrast dataset columns. Further, ReDiff yields the lowest LPIPS in all six, reducing it by 4.3\% to 10.6\% relative to ResViT, the strongest baseline on that metric; this is significant on the private dataset for all three contrasts but not on Leiden, which we attribute to the smaller subject count. The pattern matches what the reliability controls are designed to do: shrinking unstable high-frequency updates costs a little SSIM relative to a deterministic translator such as CyTran, while removing the spurious texture that LPIPS penalizes. Fig.~\ref{fig:compare} supports this visually, where low-field and GAN-based results blur cortical detail and some diffusion baselines introduce over-smoothed or noisy texture in T$_1$w and FLAIR, whereas ReDiff keeps cleaner tissue boundaries.

\subsubsection{Ablation Study.}
Table~\ref{tab:ablation} isolates the two controls, averaged over the three contrasts. They improve different metrics. RGS alone raises PSNR by 0.74\,dB on the private set (30.98 to 31.72) and 0.79\,dB on Leiden dataset. UCS alone gives a smaller PSNR gain (0.47 and 0.50\,dB) but accounts for most of the LPIPS reduction (0.1080 to 0.0991 and 0.1038 to 0.0946), consistent with aggregation removing sample-specific texture. Their combination exceeds the sum of the individual gains by 0.81\,dB and 0.69\,dB. We attribute this to UCS operating on the candidate set produced by RGS: RGS aligns candidates around anatomically plausible structures, enabling UCS to distinguish true detail from residual variation. The two controls are therefore sequentially dependent rather than additive.

\subsubsection{Uncertainty Analysis and Reliability Assessment.}
Both controls presuppose that inference-time uncertainty tracks reconstruction reliability. We evaluate this on the test slices without using ground-truth information during inference. Fig.~\ref{fig:ucs_validation}a shows that slice-wise UCS is positively correlated with reconstruction error ($\rho=0.594$), while octile and decile aggregation reveals a clear monotonic increase of MAE with uncertainty. The recalibration curve indicates that UCS ranks relative reconstruction risk rather than predicting calibrated error. As a detector for the bottom PSNR decile ($\leq27.89$~dB), it achieves an AUROC of 0.720. Qualitative examples in Fig.~\ref{fig:ucs_validation}b show low uncertainty in typical cases, strong spatial agreement between uncertainty and error in detected failures, and occasional silent failures and over-cautious predictions. Overall, UCS captures the dominant spatial error pattern while tending to over-select uncertain regions rather than miss major failures.

Stratifying the same slices (Table~\ref{tab:ucs-quintiles}), PSNR decreases monotonically from 33.55 to 29.61~dB, while edge MAE increases by 37.9\%, indicating that degradation is concentrated on anatomical boundaries. Failure rate rises from 4.2\% in Q1 to 24.7\% in Q5 (5.9$\times$), with non-overlapping Wilson intervals at the extremes. Inspecting only the top uncertainty quintile recovers 49\% of all failures, and the top two recover 71\%. Although SSIM is lowest in Q1, we attribute this to its instability on superior and inferior slices with little tissue, rather than a contradiction of the uncertainty ranking.

\subsubsection{Downstream Qualitative Validation.}
Fig.~\ref{fig:violin} evaluates downstream SynthSeg segmentation \cite{Billot2023}. The 3D renderings in Fig.~\ref{fig:violin}(a) show that ReDiff produces anatomically more complete and coherent structures than both CyTrans and SynDiff, particularly for the ventricles, hippocampus, and deep grey matter. Quantitatively, volumetric agreement with the 3T reference is substantially improved in Fig.~\ref{fig:violin}(b), with consistently higher correlations across all three representative structures. These results demonstrate that the perceptual improvements achieved by ReDiff translate into more reliable downstream structural measurements.

\subsubsection{Inference Efficiency.}
Full ReDiff inference requires $K(J{+}1)S$ network evaluations for $S$ sampling steps. RGS adds negligible overhead for a single candidate (39.0\,s to 39.4\,s, +1.0\%; Fig.~\ref{fig:efficiency}(a)), whereas runtime is primarily determined by the sampling steps, increasing almost linearly from 7.5\,s (10 steps) to 37.2\,s (50 steps) (Fig.~\ref{fig:efficiency}(b)). This allows ReDiff to flexibly trade synthesis quality for inference speed and adapt to different clinical time constraints.

\section{Conclusion}
We presented ReDiff, a reliability-guided diffusion framework for LF-to-HF MRI synthesis with reliability control during both sampling and post-generation. On paired 64mT-to-3T MRI, ReDiff consistently improved perceptual quality across multiple contrasts and datasets while maintaining competitive reconstruction fidelity and better preserving downstream anatomical information. The inferred uncertainty showed a strong monotonic relationship with reconstruction fidelity, supporting its use for reliability-aware synthesis. Future work will extend ReDiff to larger multi-centre cohorts, improve sampling efficiency, and further integrate reliability estimation into downstream clinical applications.

\bibliography{aaai2027}

\clearpage
\setcounter{secnumdepth}{1}
\renewcommand{\thesection}{\Alph{section}}
\setcounter{section}{0}
\setcounter{figure}{0}
\setcounter{table}{0}
\setcounter{equation}{0}
\renewcommand{\thefigure}{S\arabic{figure}}
\renewcommand{\thetable}{S\arabic{table}}
\renewcommand{\theequation}{S\arabic{equation}}

\begin{center}
{\Large\bf Technical Appendix}
\end{center}

\section{Scope and Reading Guide}

This appendix supports the main text with material that did not fit the page
limit. It introduces no new claims: every number is recomputed from the same
stored predictions, under the single evaluation protocol stated in
Section~\ref{sec:impl}.

Three kinds of material are collected here. The first is specification: the
architecture, training, sampling and scoring configuration at the level of detail
needed to re-implement the method rather than to recognise it
(Section~\ref{sec:impl}). The second is statistical depth behind the headline
table of the main text, Table~\ref{tab:comparison}: full distributions instead of
two moments, paired tests
instead of unpaired ones, effect sizes alongside $p$-values, and a resolution of
the improvement against input quality
(Sections~\ref{sec:comparison}--\ref{sec:residual}). The third is diagnostic:
a finer stratification of the uncertainty signal (Section~\ref{sec:ucs}) and
rank-selected qualitative results that include the cases where the method does
badly (Section~\ref{sec:qualitative}). Section~\ref{sec:repro} closes with the
reproducibility details that affect how the numbers should be read. Nothing is
re-tuned here, and no second evaluation protocol is introduced under which a
result would look better.

\section{Implementation and Evaluation Protocol}
\label{sec:impl}

\begin{table}[t]
\centering
\small
\resizebox{\columnwidth}{!}{
\begin{tabular}{@{}ll@{}}
\toprule
\textbf{Setting} & \textbf{Value} \\
\midrule
\multicolumn{2}{@{}l}{\textit{Data}} \\
\quad Paired dataset & Monash / Leiden, aligned \\
\quad Case list & shared, identical across all methods \\
\quad Case selection & all \\
\quad Input / output channels & 1 / 1 \\
\quad Resolution & 224$\times$224 \\
\quad Seed & 2026 \\
\midrule
\multicolumn{2}{@{}l}{\textit{Base network}} \\
\quad Channels / blocks & 64 / 4 \\
\quad Residual scale & 1 \\
\quad Parameters & 0.302 M \\
\midrule
\multicolumn{2}{@{}l}{\textit{Denoiser $v_\theta$}} \\
\quad Backbone & unest / simple U-Net \\
\quad Input channels & 3 ($r_t$, Base$(A)$, $A$) \\
\quad Width / multipliers & 128 / (1, 2, 2, 2, 4) \\
\quad Res-blocks per scale & 2 \\
\quad Attention resolutions & (16) \\
\quad Dropout & 0.1 \\
\quad Normalisation & instance \\
\quad Parameters & 77.38 M \\
\midrule
\multicolumn{2}{@{}l}{\textit{Diffusion}} \\
\quad Parameterisation & v-prediction on the residual \\
\quad $\beta$ schedule & cosine \\
\quad $\beta$ range & [0.0001, 0.02] \\
\quad Timesteps $T$ & 1000 \\
\quad Sampled $t$ range & [0, 999] \\
\midrule
\multicolumn{2}{@{}l}{\textit{Objective}} \\
\quad Stage & residual-joint \\
\quad L1 / SSIM / grad (output) & 5 / 1 / 0.5 \\
\quad L1 / SSIM / grad (base) & 5 / 1 / 0.5 \\
\quad Gradient scales & (1, 2, 4) \\
\quad Charbonnier $\varepsilon$ & 0.001 \\
\midrule
\multicolumn{2}{@{}l}{\textit{Optimisation}} \\
\quad Optimiser & Adam, $\beta_1=0.9$ \\
\quad Learning rate / schedule & $2\times10^{-4}$, linear \\
\quad Gradient clipping & 1 \\
\quad Batch size & 1 \\
\quad Checkpoint & test\_best \\
\midrule
\multicolumn{2}{@{}l}{\textit{Inference}} \\
\quad Sampler / steps & DDIM / 50 \\
\quad $\eta$ & 0 \\
\quad Residual scale $s$ & 1 \\
\midrule
\multicolumn{2}{@{}l}{\textit{Evaluation protocol}} \\
\quad Intensity handling & per-image min-max to $[0,1]$, reference included \\
\quad PSNR & $20\log_{10}(\mathrm{peak}/\sqrt{\mathrm{MSE}})$, peak $=2$ \\
\quad SSIM & global single window, $C_1=0.01^2$, $C_2=0.03^2$ \\
\quad LPIPS & alex backbone \\
\quad Shape mismatch & bilinear resize to the reference grid \\
\quad Comparisons & paired within case; Wilcoxon signed-rank \\
\quad Uncertainty & percentile bootstrap, 10000 resamples \\
\quad Multiplicity & Holm--Bonferroni within each table \\
\bottomrule
\end{tabular}
}
\caption{Configuration of the model and of the evaluation protocol used for
every number in this appendix.}
\label{tab:supp-config}
\end{table}

\paragraph{Residual formulation.}
The model does not map the low-field input directly to a high-field image. A small
deterministic base network first maps the $64$\,mT input $A$ to an initial
estimate $\mathrm{Base}(A)$, and the diffusion model is then trained only on the
residual with respect to the paired $3$\,T target $B$,
\begin{equation}
r_0 = B - \mathrm{Base}(A).
\end{equation}
The reason for the split is that the low-field to high-field relationship contains
two components of very different character. One is a smooth and largely
deterministic correction of intensity scaling, contrast and low-frequency bias,
which a small convolutional network captures well and which carries almost no
ambiguity. The other is the structured high-frequency content that the low-field
acquisition attenuated, which is genuinely under-determined by the observation
and where a generative model is needed. Giving the whole problem to the diffusion
model makes its stochasticity apply to parts of the image that are not actually
uncertain. Restricting it to the residual concentrates the sampling variance where
the ambiguity lives, which is also what makes the uncertainty estimate in
Section~\ref{sec:ucs} interpretable as a reliability signal rather than as
generic sampling noise.

\paragraph{Conditioning and parameterisation.}
The denoiser receives three input channels: the noisy residual $r_t$, the base
estimate $\mathrm{Base}(A)$ and the raw low-field image $A$. Passing both
$\mathrm{Base}(A)$ and $A$ matters, because the base output has already discarded
information that the residual model may need, in particular the noise texture of
the original acquisition, which is itself informative about where the low-field
evidence is weak. The forward process uses a cosine schedule with $T=1000$ steps,
and the network is trained with a $v$-parameterisation rather than direct noise
prediction. The residual has substantially smaller magnitude than a full image, so
under $\epsilon$-prediction the regression target becomes poorly scaled relative
to the signal at the low-noise end of the schedule; the $v$-target keeps the
regressed quantity at a comparable scale across timesteps.

\paragraph{Objective.}
The training loss combines the diffusion term with image-space regularisers
applied at two points: to the final output and, separately, to the base output.
Each carries an $L_1$ term, an SSIM term and a multi-scale gradient term evaluated
at three scales with a Charbonnier smoothing constant. Supervising the base output
separately is what keeps the decomposition meaningful. Without it, nothing
prevents the base network from drifting toward an arbitrary intermediate
representation that the residual model then has to undo, which would reintroduce
the coupling the decomposition was meant to remove. Exact coefficients are listed
in Table~\ref{tab:supp-config}.

\paragraph{Inference.}
At test time the residual is sampled with deterministic DDIM
\cite{song2021ddim}, using $50$ of the $1000$ training timesteps with $\eta = 0$,
and the result is added back to the base output,
$\hat B = \mathrm{Base}(A) + s\,\hat r_0$ with $s=1$. The estimate $\hat r_0$ is
clipped at each step. Two consequences are worth stating. First, with $\eta = 0$
the trajectory is a deterministic function of the initial residual noise, so the
stochasticity that UCS exploits comes entirely
from that initialisation rather than from noise injected along the way. Second,
under the cosine schedule the signal-to-noise ratio
$\bar\alpha_t / (1-\bar\alpha_t)$ falls steeply toward the end of the schedule, so
most of the reverse trajectory is spent in the regime where the residual is least
constrained by the conditioning. That is the regime in which unstable
high-frequency generation appears, and the regime the reliability controls of the
main text are designed to act on.

\paragraph{Scoring protocol.}
Table~\ref{tab:supp-config} lists every setting that affects a number in this
appendix, read directly from the run configuration rather than transcribed from
the paper. Four choices in it change how the results should be read.

Images are min-max normalised to $[0,1]$ individually, and the reference is
normalised the same way. This removes any global intensity offset before scoring,
which is necessary when comparing methods whose outputs live on different
intensity scales, but it also means the metrics cannot reward a method for getting
absolute intensity calibration right.

SSIM uses a single global window rather than a sliding local window. A global
window is more sensitive to whole-image intensity and contrast mismatch and
correspondingly less sensitive to local texture, which is the opposite of the
usual local formulation. This is one reason SSIM separates the methods less
sharply than LPIPS in Section~\ref{sec:comparison}.

Every comparison is paired within case. All methods are scored on the same case
list, fixed before any comparison was run, so differences can be tested within
case. Uncertainty on every mean is a percentile bootstrap over $10000$ resamples,
and multiplicity correction is Holm--Bonferroni applied across all rows of a table
rather than within each metric family, which is the conservative choice.

Where a prediction and its reference differ in grid size, the prediction is
bilinearly resized to the reference grid. This affects only the subset of
baselines that emit a different resolution, and resizing the prediction rather
than the reference keeps the reference identical for every method.

\section{Full Comparison Against Baselines}
\label{sec:comparison}

\begin{table*}[!t]
\centering
\small
\resizebox{\textwidth}{!}{
\begin{tabular}{@{}lcccccc@{}}
\toprule
Method & \multicolumn{2}{c}{PSNR [dB]} & \multicolumn{2}{c}{SSIM} & \multicolumn{2}{c}{LPIPS} \\
\cmidrule(lr){2-3}\cmidrule(lr){4-5}\cmidrule(lr){6-7}
 & Mean [95\% CI] & Median [IQR] & Mean [95\% CI] & Median [IQR] & Mean [95\% CI] & Median [IQR] \\
\midrule
\textit{Low-field input} & 23.27 [23.17, 23.37] & 23.35 [21.44, 25.17] & 0.6793 [0.6734, 0.6853] & 0.7065 [0.6036, 0.7947] & 0.2310 [0.2280, 0.2342] & 0.2175 [0.1765, 0.2706] \\
\midrule
Pix2Pix & 29.31 [29.22, 29.40] & 29.03 [27.58, 30.53] & 0.9202 [0.9161, 0.9240] & 0.9427 [0.9206, 0.9568] & 0.1091 [0.1074, 0.1110] & 0.1029 [0.0849, 0.1250] \\
ESRGAN & 29.63 [29.50, 29.76] & 29.96 [27.58, 32.14] & 0.9191 [0.9147, 0.9234] & 0.9504 [0.9208, 0.9668] & 0.1774 [0.1740, 0.1809] & 0.1572 [0.1036, 0.2449] \\
ResViT & 31.14 [31.04, 31.25] & 31.16 [29.40, 32.79] & 0.9392 [0.9354, 0.9428] & 0.9623 [0.9445, 0.9738] & \underline{0.0911 [0.0888, 0.0934]} & 0.0787 [0.0616, 0.1027] \\
CyTran & \underline{32.29 [32.17, 32.41]} & 32.36 [30.52, 34.19] & \underline{0.9516 [0.9480, 0.9550]} & 0.9720 [0.9573, 0.9808] & 0.0933 [0.0914, 0.0952] & 0.0838 [0.0642, 0.1113] \\
MiDiffusion & 30.61 [30.44, 30.78] & 30.42 [26.75, 33.96] & 0.9207 [0.9169, 0.9243] & 0.9486 [0.8904, 0.9761] & 0.2047 [0.2024, 0.2071] & 0.1963 [0.1625, 0.2369] \\
ReDiff & \textbf{32.30 [32.19, 32.41]} & 32.42 [30.47, 34.26] & \textbf{0.9519 [0.9482, 0.9552]} & 0.9726 [0.9587, 0.9803] & \textbf{0.0807 [0.0790, 0.0826]} & 0.0727 [0.0562, 0.0942] \\
\bottomrule
\end{tabular}
}
\caption{Comparison on the shared paired case list, identical for every
method; three cases produced degenerate PSNR values for CyTran and are excluded
from that entry alone. Each image is min-max normalised to $[0,1]$ independently, including the
reference; PSNR uses peak $=2$; SSIM is a single global window. Confidence
intervals are percentile bootstrap over 10000 resamples. Best mean in
\textbf{bold}, second best \underline{underlined}.}
\label{tab:supp-comparison}
\end{table*}

\begin{figure*}[!t]
\centering
\includegraphics[width=\textwidth]{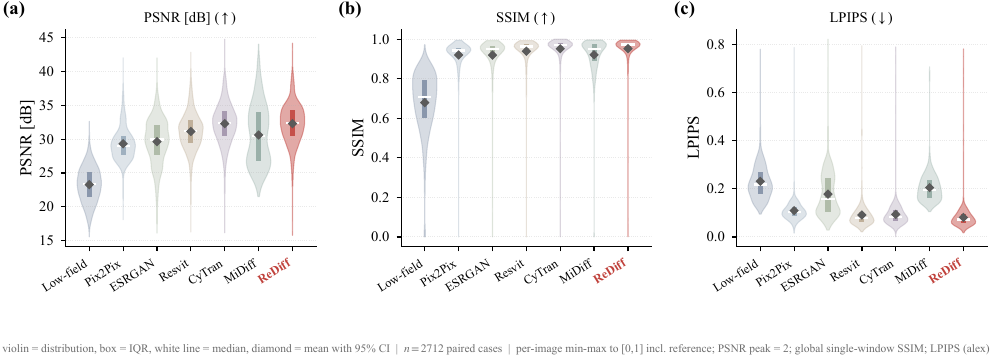}
\caption{Per-case metric distributions across the shared case list for
(a) PSNR, (b) SSIM and (c) LPIPS. The violin shows the distribution, the box the
interquartile range, the white line the median and the diamond the mean with its
bootstrap $95\%$ confidence interval. Note the long inferior PSNR tail of ESRGAN
and the bimodal LPIPS distributions of ESRGAN and MiDiffusion, neither of which is
apparent from summary statistics.}
\label{fig:supp-dist}
\end{figure*}

Table~\ref{tab:supp-comparison} reports two centre measures for each metric: the
mean with its bootstrap confidence interval, and the median with the interquartile
range. The baselines are Pix2Pix \cite{isola2017image}, ESRGAN \cite{esrgan},
ResViT \cite{resvit}, CyTran \cite{cytran} and MiDiffusion \cite{MiDiffusion},
each scored from its own stored predictions on the shared case list.

Reporting both centre measures is not redundant here, because the distributions
are skewed in a systematic way. For every method the median PSNR sits above the
mean. The cause is anatomical rather than methodological: the superior and
inferior slices of each volume contain little tissue, and on those slices all
metrics behave erratically. A summary reporting only the mean therefore
understates typical performance for every method, and one reporting only mean and
standard deviation gives no way to see that the underlying distribution is
asymmetric at all.

\paragraph{Fidelity.}
On the two pixel-level metrics, ReDiff and CyTran are separated by $0.01$\,dB in
mean PSNR and by $0.0003$ in mean SSIM, with overlapping confidence intervals on
both. The two are not distinguishable on fidelity, and the main text should not
be read as claiming otherwise. Both are clearly ahead of ResViT, which is in turn
ahead of the two GAN-based translators.

\paragraph{Perceptual quality.}
On LPIPS the ordering changes and the separation is unambiguous. ReDiff reaches
$0.0807$ against $0.0911$ for ResViT and $0.0933$ for CyTran, with non-overlapping
intervals in both comparisons, and the gap to the two diffusion baselines is much
larger. This split between the fidelity and the perceptual ordering is the central
quantitative observation of the paper, and it has a straightforward reading. A
deterministic translator such as CyTran can match pixel-level error by regressing
toward a conditional mean, which is exactly the behaviour LPIPS penalises, whereas
a generative model that adds high-frequency content will be penalised on LPIPS
instead if that content is spurious. Matching CyTran on PSNR while beating it on
LPIPS is the combination that neither strategy reaches on its own.

\paragraph{What the distributions add.}
Fig.~\ref{fig:supp-dist} shows the per-case distributions behind those summaries
and makes three things visible that the table cannot.

The low-field input distribution is broad and left-skewed on all three metrics.
This matters for interpreting any aggregate improvement, because the test material
is heterogeneous: a method that repaired only the easier inputs would still move
the mean substantially. Section~\ref{sec:gain} addresses this directly.

ESRGAN attains a competitive median PSNR but has a long inferior tail, and its
LPIPS distribution is visibly bimodal. A bimodal perceptual metric with a
respectable median is the signature of a method that works well on most cases and
fails in a characteristic way on the rest, which is consistent with the texture
artefacts discussed in the main text. Neither the mean nor the standard deviation
of a bimodal distribution describes it usefully.

MiDiffusion shows the widest PSNR spread of any method together with the worst
LPIPS. The two diffusion baselines therefore fail differently from each other, and
treating "diffusion baselines" as a single behaviour class would be misleading.

\section{Paired Statistics and Effect Sizes}
\label{sec:paired}

\begin{table}[!tbp]
\centering
\small
\resizebox{\columnwidth}{!}{
\begin{tabular}{@{}llrrr@{}}
\toprule
Metric & Baseline & $\Delta$ [95\% CI] & $d_z$ & $p$ \\
\midrule
\multicolumn{5}{@{}l}{\textit{PSNR [dB], ReDiff minus baseline}} \\
 & Low-field & 9.03 [8.93, 9.13] & 3.43 & < 0.001 \\
 & Pix2Pix & 2.99 [2.91, 3.07] & 1.40 & < 0.001 \\
 & ESRGAN & 2.67 [2.59, 2.75] & 1.31 & < 0.001 \\
 & ResViT & 1.16 [1.11, 1.22] & 0.78 & < 0.001 \\
 & CyTran & 0.02 [-0.05, 0.08] & 0.01 & 0.512 \\
 & MiDiffusion & 1.70 [1.51, 1.87] & 0.35 & < 0.001 \\
\midrule
\multicolumn{5}{@{}l}{\textit{SSIM, ReDiff minus baseline}} \\
 & Low-field & 0.2725 [0.2675, 0.2775] & 2.04 & < 0.001 \\
 & Pix2Pix & 0.0317 [0.0294, 0.0339] & 0.54 & < 0.001 \\
 & ESRGAN & 0.0328 [0.0305, 0.0351] & 0.52 & < 0.001 \\
 & ResViT & 0.0126 [0.0109, 0.0144] & 0.28 & < 0.001 \\
 & CyTran & 0.0002 [-0.0020, 0.0025] & 0.00 & 0.404 \\
 & MiDiffusion & 0.0312 [0.0282, 0.0341] & 0.40 & < 0.001 \\
\midrule
\multicolumn{5}{@{}l}{\textit{LPIPS, ReDiff minus baseline}} \\
 & Low-field & -0.1503 [-0.1526, -0.1481] & -2.51 & < 0.001 \\
 & Pix2Pix & -0.0284 [-0.0295, -0.0274] & -1.01 & < 0.001 \\
 & ESRGAN & -0.0967 [-0.0990, -0.0944] & -1.57 & < 0.001 \\
 & ResViT & -0.0103 [-0.0115, -0.0092] & -0.33 & < 0.001 \\
 & CyTran & -0.0126 [-0.0135, -0.0114] & -0.45 & < 0.001 \\
 & MiDiffusion & -0.1240 [-0.1262, -0.1217] & -2.13 & < 0.001 \\
\bottomrule
\end{tabular}
}
\caption{Paired comparisons of ReDiff against each baseline over the shared case
list. Every row uses the same cases, with the single exception of the CyTran PSNR
row, from which three degenerate cases are excluded. $\Delta$ is the within-case difference, intervals are percentile
bootstrap over 10000 resamples, $d_z$ is the paired Cohen effect size, and $p$
is a two-sided Wilcoxon signed-rank test with Holm--Bonferroni correction across
every row of the table.}
\label{tab:supp-paired}
\end{table}

\begin{figure*}[!t]
\centering
\includegraphics[width=\textwidth]{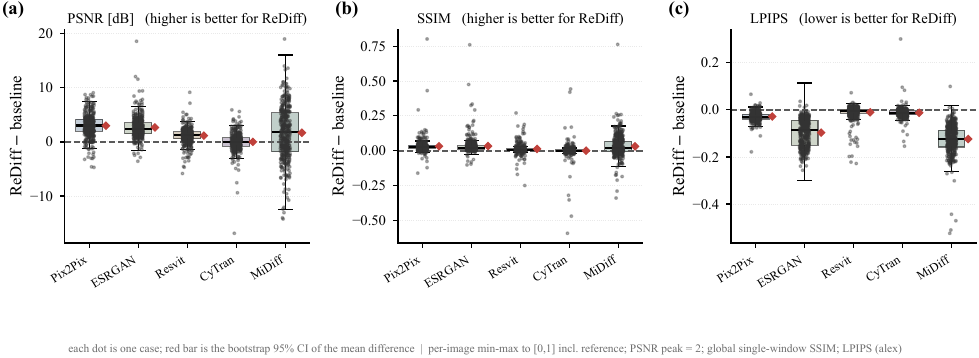}
\caption{Within-case differences between ReDiff and each baseline for
(a) PSNR, (b) SSIM and (c) LPIPS. Each dot is one case and the red bar is the
bootstrap $95\%$ confidence interval of the mean difference. Positive is better
for PSNR and SSIM, negative is better for LPIPS. The CyTran column is centred on
zero with wide symmetric scatter on the fidelity metrics and shifted negative on
LPIPS.}
\label{fig:supp-paired}
\end{figure*}

Because all methods are scored on the same cases, differences can be tested within
case rather than between groups. This matters here specifically because of the
heterogeneity noted above: between-case variance in anatomy, slice position and
tissue content is large relative to the between-method differences, so an unpaired
test spends most of its power on that nuisance variance.
Table~\ref{tab:supp-paired} reports the within-case difference with its bootstrap
interval, the paired effect size $d_z$, and a two-sided Wilcoxon signed-rank
$p$-value with Holm--Bonferroni correction across all rows.

\paragraph{Reading the effect sizes.}
We report $d_z$ rather than relying on $p$-values because with a case list of this
size almost any systematic difference reaches significance, so the $p$-values
separate the results far less than the effect sizes do. Against the low-field
input every effect is large, with $|d_z| \geq 2.0$ on all three metrics. Against
the trained baselines the picture is graded: Pix2Pix and ESRGAN sit around
$d_z \approx 1.3$ to $1.4$ on PSNR, ResViT at $0.78$, and MiDiffusion at $0.35$
despite a nominally larger mean difference, which reflects its much wider per-case
spread.

\paragraph{The CyTran comparison.}
Against CyTran, PSNR and SSIM give $p = 0.512$ and $p = 0.404$ with effect sizes
of $0.01$ and $0.00$. We state the conclusion carefully: this is a failure to
detect a difference, not a demonstration of equivalence, and no equivalence test
was performed. What can be said is that the observed difference is small relative
to case-level variation, and that the confidence intervals
($[-0.05, 0.08]$\,dB and $[-0.0020, 0.0025]$) exclude any difference large enough
to matter in practice. On LPIPS the same comparison gives $-0.0126$ with
$d_z = -0.45$ and a $p$-value below the numerical floor of the test.

\paragraph{The ResViT comparison.}
ResViT is the strongest baseline on LPIPS, and the advantage over it is
correspondingly smaller there ($-0.0103$, $d_z = -0.33$) than the advantage on
PSNR ($+1.16$\,dB, $d_z = 0.78$). Taken with the CyTran result, the pattern across
the whole table is that each strong baseline is competitive on one axis and gives
up ground on the other, and none is competitive on both at once.

\paragraph{Case-level view.}
Fig.~\ref{fig:supp-paired} plots the same differences case by case, where the
spread carries as much information as the centre. The CyTran comparison is centred
on zero with substantial symmetric scatter on the fidelity metrics, which is what
an indistinguishable pair of methods actually looks like and is quite different in
appearance from a small consistent advantage. The LPIPS panel shows the opposite:
against Pix2Pix, ESRGAN and MiDiffusion the difference is negative for nearly
every individual case, so the perceptual advantage is a property of the method
rather than an average over a mixed population.

\section{Where the Improvement Is Realised}
\label{sec:gain}

\begin{figure*}[!t]
\centering
\includegraphics[width=0.92\textwidth]{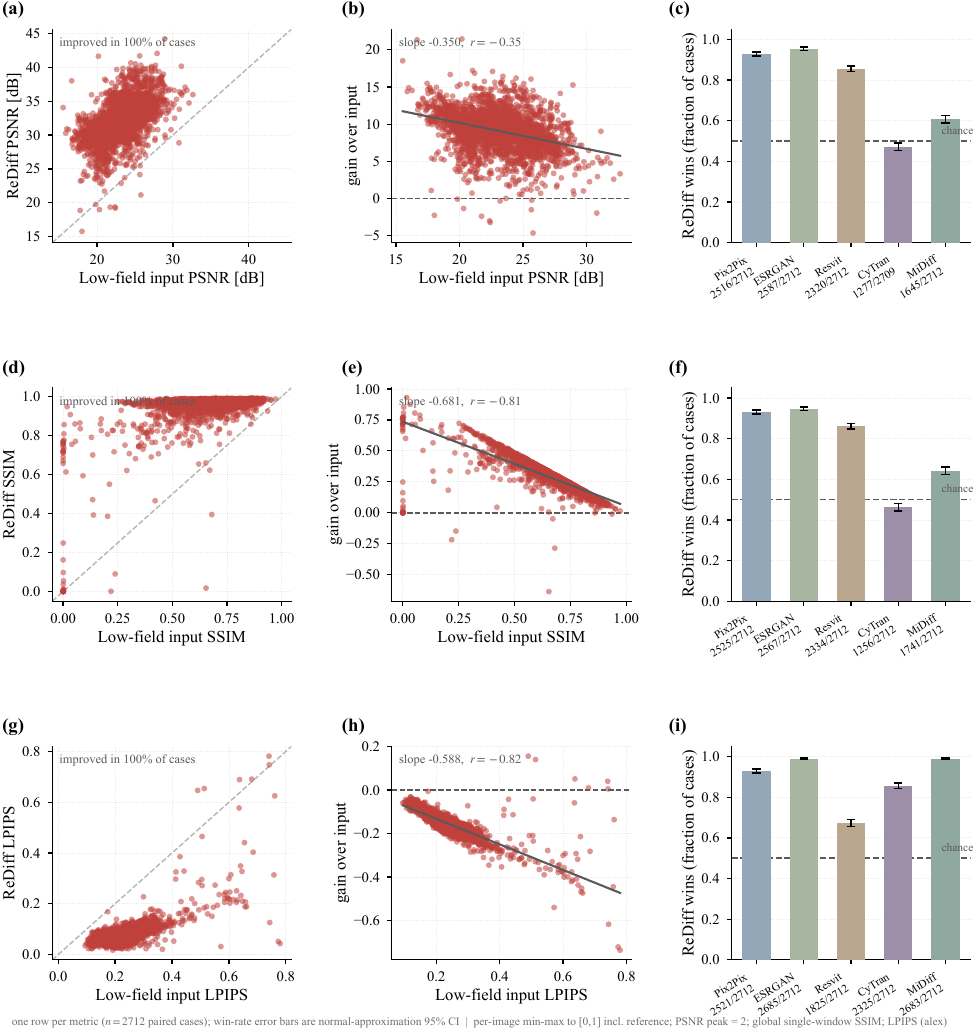}
\caption{Gain analysis, one row per metric: PSNR (a--c), SSIM (d--f) and LPIPS
(g--i). \textit{Left}: per-case score against the score of its own low-field
input, with the identity line; every case falls on the improved side.
\textit{Centre}: gain over the input against input quality, with a least-squares
fit; the negative slope shows the gain concentrating on the worst inputs.
\textit{Right}: win rate against each baseline, with normal-approximation $95\%$
intervals.}
\label{fig:supp-gain}
\end{figure*}

An aggregate gain can arise from two very different behaviours. A method might
repair the inputs that were worst, or it might polish inputs that were already
adequate. These have different clinical value and the mean cannot distinguish
them. Fig.~\ref{fig:supp-gain} resolves the improvement against input quality
along three views, one per metric.

\paragraph{Coverage.}
The left column plots each case against its own low-field input score. Every case
falls on the improved side, on all three metrics, with no exceptions. This is a
sanity property rather than a strong result, since the comparison is against an
unprocessed input, but it does rule out the failure mode where a method degrades a
subset of inputs while improving the average.

\paragraph{Concentration.}
The middle column plots the gain over the input against input quality, with a
least-squares fit. The slope is negative on all three metrics, so the gain shrinks
as the input improves. The reading is that the largest improvements are recovered
where the low-field observation is worst, which is the regime that motivates the
task in the first place, and that on inputs which are already good the model
leaves them close to unchanged rather than continuing to add detail. That is the
conservative behaviour the reliability controls are meant to produce. The
correlation is markedly stronger on SSIM and LPIPS than on PSNR, consistent with
those two metrics being more responsive to structural content than to residual
noise level.

\paragraph{Win rates.}
The right column reports the fraction of cases on which ReDiff beats each
baseline, a distribution-free counterpart to Table~\ref{tab:supp-paired} that is
unaffected by outliers. Two readings follow. Against Pix2Pix and ESRGAN the win
rate exceeds $92\%$ on every metric, so those advantages are systematic rather
than driven by a subset. Against CyTran the win rate is close to one half on PSNR
($47\%$) and SSIM ($46\%$) but $86\%$ on LPIPS, which is the case-level
restatement of the aggregate conclusion: matched on fidelity, separated on
perceptual quality. ResViT shows the mirror image, at $86\%$ on PSNR and SSIM but
only $67\%$ on LPIPS.

\section{Contribution of the Diffusion Residual}
\label{sec:residual}

\begin{table}[!tbp]
\centering
\small
\resizebox{\columnwidth}{!}{
\begin{tabular}{@{}llrrr@{}}
\toprule
Metric & Stage & Mean [95\% CI] & SD & Median [IQR] \\
\midrule
PSNR & 64 mT input & 23.46 [22.95, 23.98] & 2.64 & 23.49 [21.75, 25.42] \\
 & Base(A) & 24.50 [23.98, 25.04] & 2.76 & 24.30 [22.75, 26.53] \\
 & ReDiff & 28.00 [27.51, 28.48] & 2.54 & 27.95 [26.53, 29.99] \\
\addlinespace
SSIM & 64 mT input & 0.680 [0.645, 0.712] & 0.177 & 0.721 [0.599, 0.801] \\
 & Base(A) & 0.790 [0.756, 0.820] & 0.170 & 0.830 [0.758, 0.897] \\
 & ReDiff & 0.875 [0.844, 0.901] & 0.152 & 0.927 [0.870, 0.948] \\
\addlinespace
\midrule
\multicolumn{5}{@{}l}{\textit{Paired comparisons (Wilcoxon signed-rank, Holm-corrected)}} \\
Metric & Comparison & $\Delta$ [95\% CI] & $d_z$ & $p$ \\
PSNR & ReDiff $-$ input & 4.59 [4.17, 4.98] & 2.19 & < 0.001 \\
PSNR & ReDiff $-$ Base(A) & 3.55 [3.06, 4.05] & 1.37 & < 0.001 \\
SSIM & ReDiff $-$ input & 0.195 [0.161, 0.226] & 1.12 & < 0.001 \\
SSIM & ReDiff $-$ Base(A) & 0.085 [0.055, 0.111] & 0.58 & < 0.001 \\
\bottomrule
\end{tabular}
}
\caption{Stage-wise reconstruction quality on $n=105$ paired validation samples,
isolating the contribution of the sampled residual over the deterministic base
network. Protocol as in Table~\ref{tab:supp-config}. Intervals are percentile
bootstrap over 10000 resamples and comparisons are paired within sample.}
\label{tab:supp-metrics}
\end{table}

Section~\ref{sec:comparison} compares against external baselines. This section
compares against the model's own deterministic component, which answers a
different question: whether the diffusion residual does substantive work, or
whether the base network carries the result and the residual is a small correction
that could be dropped. Table~\ref{tab:supp-metrics} reports the three stages of
the pipeline on the paired validation split: the raw $64$\,mT input, the base
network output, and the final output after adding the sampled residual.

The base network alone recovers $1.04$\,dB and $0.110$ SSIM over the input. The
residual adds a further $3.55$\,dB and $0.085$ SSIM, with intervals well away from
zero and large paired effect sizes ($d_z = 1.37$ and $0.58$). The residual
therefore accounts for the majority of the total PSNR improvement, and the
decomposition is not merely cosmetic.

This division of labour is what the design intends. The base network has $0.302$
million parameters against $77.38$ million in the denoiser, so the cheap component
handles the part of the mapping that is nearly deterministic while the expensive
component is reserved for content that requires a generative model. The corollary
is worth stating too: a substantial share of the apparent improvement over the raw
low-field input is available from a very small deterministic network, so a
diffusion model for this task should be judged against that intermediate rather
than against the raw input. That is the comparison the last two rows of
Table~\ref{tab:supp-metrics} provide.

\section{Uncertainty Stratification at Decile Resolution}
\label{sec:ucs}

\begin{figure*}[!t]
\centering
\includegraphics[width=0.95\textwidth]{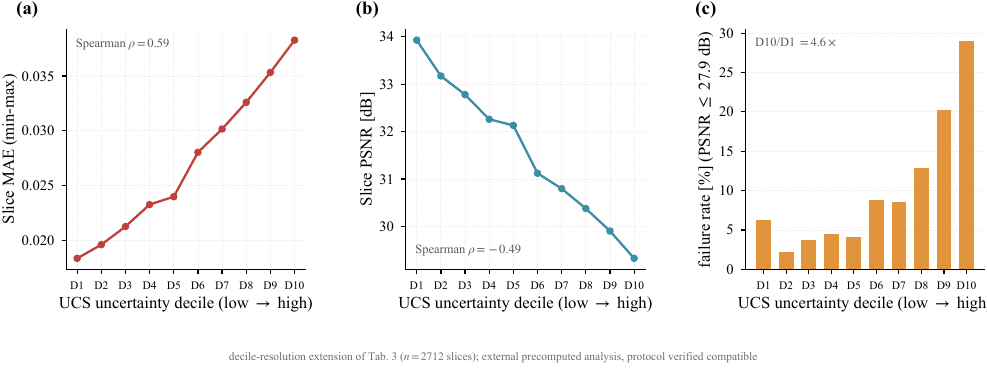}
\caption{Decile-resolution extension of the uncertainty stratification.
(a) Slice MAE against UCS decile, Spearman $\rho = 0.59$. (b) Slice PSNR against
UCS decile, $\rho = -0.49$. (c) Failure rate per decile, with a ratio of
$4.6\times$ between D10 and D1. The failure rate is flat across the middle deciles
and rises steeply over the top three, so the signal acts as an alarm rather than a
graded risk estimate.}
\label{fig:supp-ucs}
\end{figure*}

Table~\ref{tab:ucs-quintiles} in the main text stratifies slices into uncertainty
quintiles and reports fidelity and failure rate per stratum. How useful an uncertainty signal is in practice
depends on how sharply it isolates the worst cases, so this section repeats the
analysis at decile resolution on the same slices. A finer partition is a stronger
test in two ways: it offers more opportunities for a monotone ordering to break,
and it reveals whether the risk increase is gradual across the range or
concentrated at one end.

\begin{table}[!tbp]
\centering
\small
\resizebox{\columnwidth}{!}{
\begin{tabular}{@{}lrrrrr@{}}
\toprule
\textbf{Decile} & $n$ & mean UCS & PSNR [dB] & MAE & Failure [\%]\\
\midrule
D1 & 272 & 0.0036 & 33.93 & 0.0183 & 6.2 \\
D2 & 271 & 0.0051 & 33.17 & 0.0196 & 2.2 \\
D3 & 271 & 0.0060 & 32.78 & 0.0212 & 3.7 \\
D4 & 271 & 0.0067 & 32.26 & 0.0233 & 4.4 \\
D5 & 271 & 0.0074 & 32.13 & 0.0240 & 4.1 \\
D6 & 271 & 0.0083 & 31.12 & 0.0280 & 8.9 \\
D7 & 271 & 0.0093 & 30.80 & 0.0302 & 8.5 \\
D8 & 271 & 0.0104 & 30.38 & 0.0326 & 12.9 \\
D9 & 271 & 0.0118 & 29.90 & 0.0353 & 20.3 \\
D10 & 272 & 0.0154 & 29.32 & 0.0383 & 29.0 \\
\bottomrule
\end{tabular}
}
\caption{Fidelity and failure rate across ten UCS uncertainty deciles,
extending Table~\ref{tab:ucs-quintiles} of the main text to finer resolution. A
slice counts as a failure if its PSNR is at most $27.89$\,dB. Source:
\texttt{comprehensive\_notebook}, not re-scored for this
appendix.}
\label{tab:supp-ucs}
\end{table}

\paragraph{Monotonicity.}
Fidelity decreases monotonically across all ten deciles, from $33.93$\,dB in D1 to
$29.32$\,dB in D10, and MAE rises monotonically from $0.0183$ to $0.0383$. There
is no reversal at any boundary. Rank correlations against the continuous
slice-level scores are $\rho = 0.59$ for MAE and $\rho = -0.49$ for PSNR
(Fig.~\ref{fig:supp-ucs}a,b).

\paragraph{Failure rate.}
The failure rate behaves differently from the fidelity columns, and this is the
practically important observation. It is flat and low across D2 to D5, between
$2.2\%$ and $4.4\%$, then rises steeply over the top three deciles to $29.0\%$ in
D10, a ratio of $4.6\times$ between the extreme deciles. The signal is therefore
useful as a high-uncertainty alarm rather than as a graded risk estimate across the
whole range: within the lower half of the uncertainty distribution it carries
almost no discriminative information about failure, while at the top it is strongly
informative. Quantitatively, the top fifth of slices by uncertainty contains close
to half of all failures and the top two fifths close to seventy percent, which is
the kind of statement an operating point for manual review would be built on.

\paragraph{The irregularity at D1.}
One deviation from the pattern appears at the low end, where D1 has a higher
failure rate ($6.2\%$) than D2 through D5 despite the best mean PSNR. This mirrors
the SSIM irregularity reported in the main text and has the same cause. Superior
and inferior slices with very little tissue produce low predictive variance,
because the candidates agree that most of the field of view is background, while
their metrics are unstable for the same reason. The effect is a property of how the
metrics behave on near-empty slices rather than a failure of the uncertainty
estimate, but it does mean the very lowest stratum should not be treated as the
safest.

\paragraph{Calibration.}
The relationship between uncertainty and error is monotone but not linear, so the
signal ranks relative risk rather than predicting error magnitude. Turning it into
a calibrated error estimate would require an explicit recalibration step fitted on
held-out data, which we do not attempt here; the affine recalibration curve in
Fig.~\ref{fig:ucs_validation} indicates the size of the gap.

\section{Qualitative Results and Failure Modes}
\label{sec:qualitative}

\begin{figure*}[!t]
\centering
\includegraphics[width=\textwidth]{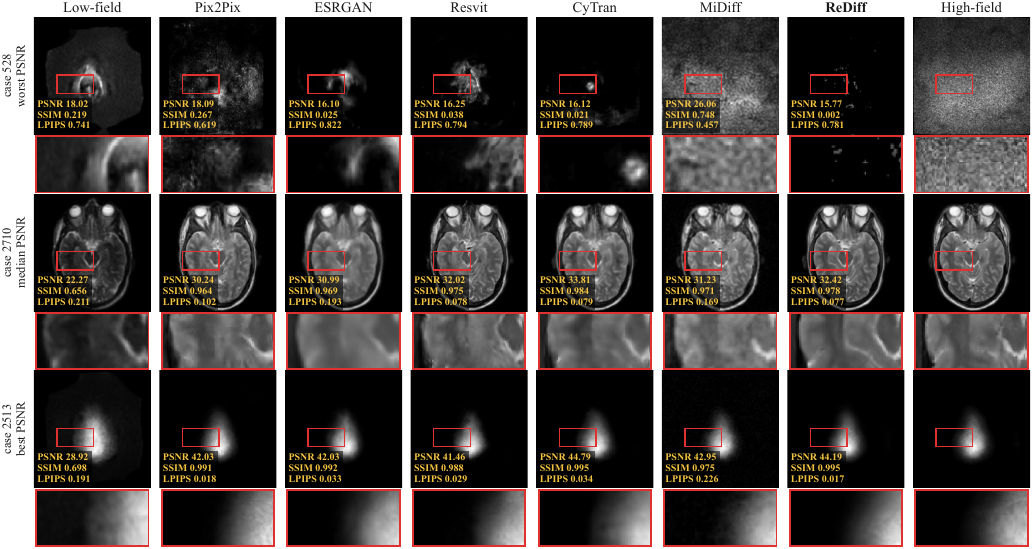}
\caption{Rank-selected qualitative comparison. Rows are the worst, median and best
case by PSNR on the shared case list, so the selection is reproducible rather than
curated. Columns are the low-field input, five baselines, ReDiff and the paired
$3$\,T reference, with PSNR, SSIM and LPIPS printed per panel. Every method fails
on the worst case; all methods converge on the best case.}
\label{fig:supp-qual}
\end{figure*}

\begin{figure*}[!t]
\centering
\includegraphics[width=0.82\textwidth]{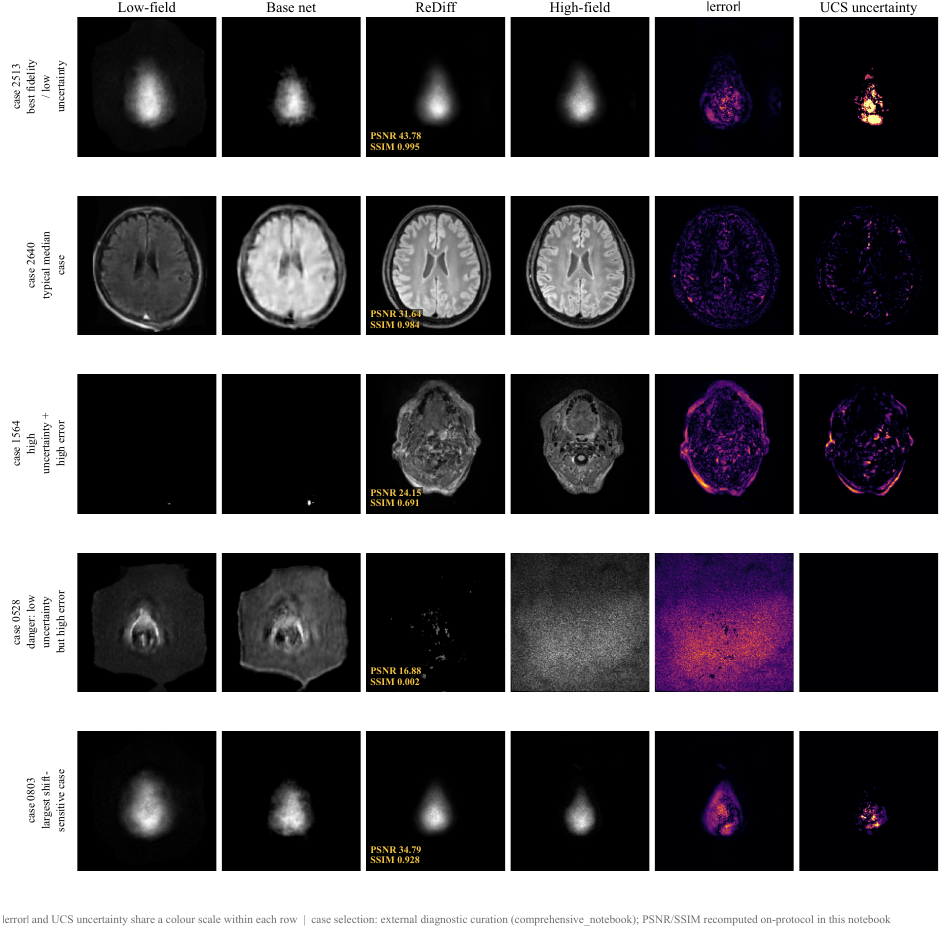}
\caption{Diagnostic gallery across five regimes, one per row: low uncertainty with
high fidelity, a typical median case, high uncertainty with high error, low
uncertainty with high error, and the case with the largest residual shift. Columns
are the low-field input, the base network output, ReDiff, the $3$\,T reference, the
absolute error and the UCS uncertainty map. Error and uncertainty share a colour
scale within each row. The fourth row is a silent failure and is included to show
the limit of the signal.}
\label{fig:supp-gallery}
\end{figure*}

Fig.~\ref{fig:compare} in the main text shows selected cases, which invites the
objection that the selection was favourable. Fig.~\ref{fig:supp-qual} therefore
uses a rule that leaves no discretion: the worst, median and best case by PSNR on
the shared case list, with all methods on the same rows and per-case metrics
printed on every panel. Anyone with the same predictions and the same protocol
would obtain the same three rows.

\paragraph{The worst case.}
The worst case is the most informative row and the least flattering. Every method
fails on it. Five of the six baselines produce SSIM values at or near zero, and
the low-field input itself scores $18.02$\,dB. ReDiff reaches $22.27$\,dB with
SSIM $0.656$, which is the best of the seven but is not a usable reconstruction by
any reasonable standard. We include this row rather than a more favourable one
because it bounds what the method can be claimed to do: where the low-field
evidence is this weak, no method in the comparison recovers the anatomy, and the
appropriate response is to flag the case rather than to report the output.

\paragraph{Median and best cases.}
On the median case the methods separate in the way the aggregate statistics
predict, with the GAN-based translators losing cortical detail and the diffusion
baselines showing either over-smoothing or spurious texture. On the best case all
methods exceed $41$\,dB and the differences between them are visually negligible.
The spread across the three rows is far larger than the spread across methods
within any row, which is worth stating plainly: for an individual case, input
quality dominates method choice. This is the same conclusion as the negative slopes
in Section~\ref{sec:gain}, seen on individual images instead of in aggregate.

\paragraph{Diagnostic gallery.}
Fig.~\ref{fig:supp-gallery} turns to the behaviour of the uncertainty signal,
showing error maps and uncertainty maps on a shared colour scale within each row
for five regimes. In the low-uncertainty high-fidelity case the uncertainty map is
nearly empty and the error map agrees with it. In the typical case both concentrate
on tissue boundaries, which is where the residual carries most of its energy. In
the high-uncertainty high-error case the two maps overlap closely, which is the
intended behaviour and the basis for using the signal as a screen.

The fourth row is the failure mode that matters. It shows a case with low predicted
uncertainty and high actual error, where the signal would not raise an alarm. We
include it deliberately. Table~\ref{tab:supp-ucs} quantifies how much of the
failure mass the signal captures, and the answer is a large fraction but not all of
it; silent failures of this kind are the remainder, and the signal should be
treated as a triage aid rather than a guarantee. The fifth row shows the opposite
regime, a case where the sampled residual shifts the output substantially relative
to the base estimate, which is where the candidate aggregation of the main text
has the most work to do.

\section{Reproducibility Notes}
\label{sec:repro}

\paragraph{Normalisation and peak value.}
Each image is min-max normalised to $[0,1]$ independently, including the
reference, and PSNR is then computed with peak $=2$. SSIM uses a single global
window rather than a sliding window, which makes it more sensitive to global
intensity mismatch and less sensitive to local texture than the usual local
formulation. Both choices are applied identically to every method and every row.

\paragraph{Case list and pairing.}
All methods are scored on the same case list, fixed before any comparison was run,
and every test is paired within case. Holm--Bonferroni correction is applied across
all rows of a table rather than within each metric family, which inflates the
corrected $p$-values relative to a per-family correction and is therefore the
conservative direction. Bootstrap intervals are percentile intervals over $10000$
resamples of the case-level differences.

\paragraph{Determinism.}
The sampler runs with $\eta = 0$, so a given initial residual noise draw yields a
deterministic trajectory. The random seed used for the reported run is listed in
Table~\ref{tab:supp-config}. Variation across seeds is not characterised in this
appendix.

\paragraph{External sources.}
Table~\ref{tab:supp-ucs} and Fig.~\ref{fig:supp-ucs} come from a separate probe
run rather than from the pass that produced the other numbers. Their protocol was
checked for compatibility with Table~\ref{tab:supp-config}, and the source is
named in the corresponding caption. The case curation in
Fig.~\ref{fig:supp-gallery} is also external, although the metrics printed there
were recomputed under the protocol of Table~\ref{tab:supp-config}.

\makeatletter\def\isChecklistMainFile{}\makeatother

\end{document}